%% file: main.tex
\documentclass{article}

\usepackage{iclr2027_conference,times}
\iclrfinalcopy  

\usepackage{url}
\usepackage{latexsym}
\usepackage[T1]{fontenc}
\usepackage[utf8]{inputenc}
\usepackage{microtype}
\usepackage{inconsolata}
\usepackage{graphicx}
\usepackage{amsmath,amssymb,amsthm}
\usepackage{booktabs,multirow}
\usepackage{algorithm}
\usepackage{algpseudocode}
\usepackage{enumitem}
\usepackage{wrapfig}
\usepackage{caption}
\newtheorem{lemma}{Lemma}
\newtheorem{proposition}{Proposition}
\newtheorem{theorem}{Theorem}
\newtheorem{corollary}{Corollary}
\newtheorem*{restatement}{Restatement}
\theoremstyle{remark}
\newtheorem{remark}{Remark}

\newcommand{\our}{Muon-RT}

\newcommand{\base}{SDFT}
\newcommand{\sdpo}{SDPO}
\newcommand{\ns}{Newton--Schulz}

\newcommand{\best}[1]{\textbf{#1}}
\newcommand{\second}[1]{\underline{#1}}
\newcommand{\pub}{\textsuperscript{\dag}}

\usepackage{xcolor}
\definecolor{lightblue}{RGB}{242,248,255}
\usepackage[hidelinks]{hyperref}

\title{\our: Reusing the Teacher's Trajectory for Momentum-Free Muon for LLM Self-Distillation}
\author{
Aryan Das$^{1}$ \quad Vinay Kumar Verma$^{1}$ \quad Aakansha Mishra$^{2}$ \\
$^{1}$IIT Kanpur \quad $^{2}$Amazon
}

\begin{document}

\maketitle

\input{sections/0_abstract}
\input{sections/1_introduction}
\input{sections/2_preliminaries}
\input{sections/3_methodology}
\input{sections/3.1_theorem}
\input{sections/4_experiments}

\input{sections/2_related_work}
\input{sections/5_conclusion}

\input{sections/6_limitations}
\input{sections/7_ethics}

\bibliographystyle{iclr2027_conference}
\bibliography{main}

\clearpage
\appendix
\section{Appendix}
\label{app:main}
\input{sections/appen_proof}

\input{sections/appendix}

\end{document}

%% file: sections/0_abstract.tex
\begin{abstract}

Self-distillation methods for large language models with an exponential moving average (EMA) teacher already retain a history of student updates. Muon nevertheless stores a separate momentum buffer, increasing persistent training memory. We propose Muon with Reused Teacher (Muon-RT), which reuses the teacher–student displacement to eliminate momentum buffers for hidden weight matrices. We show that this displacement encodes an exponentially weighted history of applied updates and smooths temporally uncorrelated noise under a local model. Muon-RT removes its radial component, normalizes it spectrally, and mixes it with the current gradient after Newton–Schulz (NS) orthogonalization. Under explicit regularity and alignment assumptions, we establish linear contraction of an expected objective-gap bound toward a residual error level. With Self-Distillation Fine-Tuning (SDFT) on Qwen2.5-7B-Instruct, Muon-RT improves average accuracy over AdamW by 2.6 points with 42\% less persistent memory, and over momentum Muon by 5.4 points with 27\% less persistent memory. With Self-Distillation Policy Optimization (SDPO) on Olmo-3-7B-Instruct, it improves average accuracy over AdamW by 2.0 points.

\end{abstract}

%% file: sections/1_introduction.tex
\section{Introduction}
\label{sec:introduction}



\begin{wrapfigure}{r}{0.48\textwidth}
  \centering
  \vspace{-5mm}
  \includegraphics[width=\linewidth]{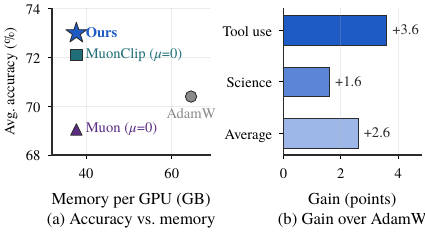}
  \vspace{-5mm}
  \caption{\our{} with \base{} on Qwen2.5-7B-Instruct. (a) Average accuracy over 2 datasets against memory per GPU. (b) Gain over AdamW in points.}
  \label{fig:teaser}
  \vspace{-3mm}
\end{wrapfigure}
Self-distillation~\citep{shenfeld2026sdft,hubotter2026sdpo} provides an effective approach to adapting large language models (LLMs) by using privileged information to elicit supervision from the model's existing capabilities. The student learns from its own generated responses by matching a teacher's predictions conditioned on additional context. This formulation supports skill acquisition and policy improvement while helping preserve previously learned capabilities.
Such retention is an advantage over conventional fine-tuning, which can cause forgetting~\citep{biderman2024lora,zheng2025spurious}, a recognized challenge when updating LLMs~\citep{shi2025clsurvey,zheng2025lifelong}. Within this family, Self-Distillation Fine-Tuning (\base{}) uses demonstrations to guide the acquisition of new skills~\citep{shenfeld2026sdft}, while Self-Distillation Policy Optimization (\sdpo{}) extends the approach to on-policy post-training using feedback from successful attempts~\citep{hubotter2026sdpo}. In this settings, the teacher is maintained as an exponential moving average (EMA) of student weights, following approaches developed in semi-supervised and self-supervised learning~\citep{tarvainen2017meanteacher,grill2020byol,caron2021dino}.
The teacher therefore evolves alongside the student, providing an averaged model whose predictions guide subsequent learning.

Optimizing EMA-based self-distillation incurs substantial memory costs because training must retain student and teacher parameters alongside optimizer state. Adam~\cite{kingma2014adam} and AdamW~\citep{loshchilov2019adamw} store two running moment estimates per parameter. Muon~\citep{jordan2024muon} reduces this overhead by maintaining one momentum buffer per hidden weight matrix and approximately orthogonalizing update directions using \ns{} iterations. Muon and its variants have been used to train Kimi~K2~\citep{kimi2025k2} and GLM-4.5~\citep{5team2025glm45agenticreasoningcoding}. Nevertheless, optimizing self-distillation with Muon still requires three persistent states for each hidden matrix: student weights, teacher weights, and momentum. AdamW requires four, including its two moment buffers. For example Qwen2.5-7B-Instruct, momentum Muon and AdamW require 103.2\,GB and 129.0\,GB of GPU memory for persistent states, respectively, compared with 75.4\,GB for \our{} (Figure~\ref{fig:teaser}). Existing memory-efficient optimizers address state storage through low-rank projection, quantization, or shared statistics~\citep{zhao2024galore,li2023fourbit,zhang2025adammini,zhu2025apollo}. The teacher offers another opportunity: reusing history already maintained by self-distillation to remove a separate optimizer buffer. We therefore ask:

\begin{center}
\setlength{\fboxsep}{8pt}
\fcolorbox{black!35}{lightblue}{%
\parbox{0.90\linewidth}{\centering\itshape
Can the teacher's existing parameter trajectory replace Muon's momentum buffer reducing persistent memory while improving task accuracy in self-distillation?}}
\end{center}



\textbf{Why could the teacher replace momentum?}
In \base{} and \sdpo{}, the teacher is not fixed; its weights continuously follow the evolving student through an EMA. Unlike momentum, which averages gradients, the teacher averages weights. The momentum-like information therefore lies in the \emph{teacher--student displacement}, not the teacher itself. As the student progresses, this displacement records an exponentially weighted history of applied updates (Lemma~\ref{lem:ema_history}), smoothing noisy steps while preserving persistent directions (Lemma~\ref{lem:ema_variance}). Under gradient descent, this history is proportional to gradient momentum; under Muon, it captures previously orthogonalized updates. As optimization settles, the displacement shrinks as the teacher catches up with the student. Although this cannot recover AdamW's second-moment statistics, however, it can replace Muon's momentum buffer.

Even though this approach appears simple, reusing the teacher--student displacement in Muon is not straightforward. First, Muon orthogonalizes its updates, so the displacement averages past optimizer outputs rather than raw gradients. Using it directly as momentum would feed these outputs back into the optimizer. Second, mixing the displacement with gradients before \ns{} leaves its influence uncontrolled, as \ns{} can amplify weak singular directions. This distinction matters in self-distillation, where the teacher and learning targets continuously evolve with the student. While momentum averages gradients from outdated targets, the displacement captures the updates actually applied to the student, providing a different source of historical guidance.

We therefore propose Muon with Reused Teacher (\our{}), which replaces the dedicated momentum buffer with information from the existing EMA teacher. For each hidden weight matrix, \our{} orthogonalizes the current gradient without momentum and separately computes the teacher--student displacement. It removes the component parallel to the student weights and scales the remainder by an estimate of its largest singular value.
The two directions are then combined after orthogonalization using a small fixed mixing coefficient. This preserves current gradient information while adding historical guidance without changing the training objective. Only one small vector per matrix is retained to warm-start power iteration across successive optimization steps. The method applies wherever a parameter-aligned EMA teacher is already maintained, including the studied \base{} and \sdpo{} settings and mean-teacher training~\citep{tarvainen2017meanteacher}.
Independent teachers lacking this alignment fall outside its scope. Our contributions are:

\begin{itemize}[leftmargin=*,itemsep=1pt,topsep=2pt]
    \item We establish that the teacher--student displacement stores a discounted history of applied updates, is proportional to gradient momentum under gradient descent, and reduces variance under a local model with a fixed direction and temporally uncorrelated noise.

    \item We introduce \our{}, eliminating hidden-matrix momentum buffers through normalized teacher--student displacement added after orthogonalization.
    Under regularity, bounded objective drift, and alignment assumptions, we establish linear contraction of an expected objective-gap bound toward a residual error level.

    \item With \base{}, \our{} improves average accuracy over AdamW by 2.6 points with 42\% less persistent memory, and over momentum Muon by 5.4 points with 27\% less (Figure~\ref{fig:teaser}). With \sdpo{}, it gains 2.0 points over AdamW. \our{} ranks first among Muon variants across both hosts' six settings.
\end{itemize}

%% file: sections/2_preliminaries.tex
\section{Preliminaries}
\label{sec:preliminaries}


\paragraph{Self-distillation with an EMA-teacher:}
Let $S$ and $T$ denote the student and teacher LLMs, with policies $\pi_S$ and $\pi_T$ parameterized by $\theta_S$ and $\theta_T$, respectively. Both share the same architecture and initialization. For a prompt $x$, the student samples a completion $y\sim\pi_S(\cdot\mid x)$. The teacher evaluates the same tokens using privileged context $c$: an expert demonstration in SDFT~\citep{shenfeld2026sdft}, or rollout feedback, potentially including a successful sibling response, in SDPO~\citep{hubotter2026sdpo}. At token position $j$, their next-token distributions are
\begin{equation}
\small
p_{S,j}=\pi_S(\cdot\mid x,y_{<j}), \quad
p_{T,j}=\pi_T(\cdot\mid x,c,y_{<j}).
\label{eq:sd_policies}
\end{equation}
The student minimizes the token-level distillation objective
\begin{equation}
\small
\mathcal{L}(\theta_S)
=
\mathbb{E}_{x,y,c}
\left[
\sum_{j=1}^{|y|}
\mathcal{D}\!\left(p_{S,j},\operatorname{sg}[p_{T,j}]\right)
\right],
\label{eq:sd_loss}
\end{equation}
where the expectation uses student-generated completions and their associated teacher contexts, $\operatorname{sg}$ stops teacher gradients, and $\mathcal{D}$ denotes the chosen divergence. SDFT formulates reverse KL but favors forward KL empirically, while SDPO employs Jensen--Shannon or reverse KL. To stabilize distillation, the EMA-teacher is updated after each student optimization step:
\begin{equation}
\small
\theta_T \leftarrow \tau\theta_T+(1-\tau)\theta_S,
\qquad \tau\in[0,1).
\label{eq:ema_update}
\end{equation}
Our method assumes only that the EMA-teacher is explicitly maintained and parameter-wise aligned with the student.

\paragraph{Muon Optimizer:}
Muon updates each student hidden weight matrix $\mathbf{W}_S\in\mathbb{R}^{A\times B}$, while embeddings, output heads, norms, and biases typically use AdamW~\citep{jordan2024muon,loshchilov2019adamw,liu2025muon}. Given gradient $\mathbf{G}=\nabla_{\mathbf{W}_S}\mathcal{L}(\theta_S)$ and momentum coefficient $\mu$, Muon maintains $\mathbf{M}\leftarrow\mu\mathbf{M}+\mathbf{G}$ and orthogonalizes the Nesterov direction:
\begin{equation}
\mathbf{O}^{\mathrm{mom}}=\mathrm{NS}_K(\mathbf{G}+\mu\mathbf{M}).
\label{eq:muon_momentum}
\end{equation}
Here, $\mathrm{NS}_K$ applies $K$ quintic Newton--Schulz iterations from $\mathbf{X}_0=\mathbf{Z}/(\|\mathbf{Z}\|_{\mathrm F}+\varepsilon)$, with $\mathbf{Z}=\mathbf{G}+\mu\mathbf{M}$:
\begin{equation}
\small
\mathbf{X}_{r+1}=\big(\kappa_1\mathbf{I}+\kappa_2\mathbf{X}_r\mathbf{X}_r^\top+\kappa_3(\mathbf{X}_r\mathbf{X}_r^\top)^2\big)\mathbf{X}_r.
\label{eq:muon_ns}
\end{equation}
The fixed coefficients approximately equalize nonzero singular values, yielding an approximate polar factor~\citep{jordan2024muon}. With decoupled weight decay, the update is
\begin{equation}
\small
\mathbf{W}_S\leftarrow(1-\eta\lambda)\mathbf{W}_S
-\eta c_{A,B}\mathbf{O}^{\mathrm{mom}}.
\label{eq:muon_update}
\end{equation}
where $\eta$ is the learning rate, $\lambda$ is weight decay, and $c_{A,B}=0.2\sqrt{\max(A,B)}$ approximately matches AdamW's update RMS~\citep{liu2025muon}. Combined with the EMA-teacher, Muon maintains 3 persistent states per hidden matrix: student weights $\mathbf{W}_S$, teacher weights $\mathbf{W}_T$, and momentum $\mathbf{M}$. We do not count the gradient $\mathbf{G}$ as a persistent state: it is recomputed at every step and freed after the update, and every optimizer needs it in the same way. During a step, Muon therefore holds 4 matrix-sized tensors, and between steps it keeps 3.


%% file: sections/3_methodology.tex




\section{Momentum-Free Muon Optimization}
\label{sec:method}

\begin{figure}[t]
  \centering
  \includegraphics[width=\linewidth]{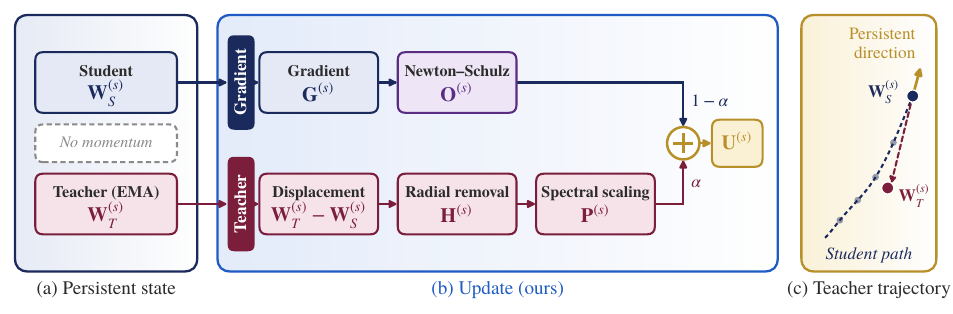}
  \vspace{-5mm}
  \caption{Overview of \our{}. (a) Each hidden matrix maintains student weights $\mathbf{W}_S$ and EMA-teacher weights $\mathbf{W}_T$, eliminating Muon's momentum buffer $\mathbf{M}$. (b) Newton--Schulz approximately orthogonalizes the gradient $\mathbf{G}$ into $\mathbf{O}$. The radial component of the teacher--student displacement $\mathbf{W}_T-\mathbf{W}_S$ is removed to obtain $\mathbf{H}$, which is spectrally normalized into $\mathbf{P}$. The update combines both directions as $\mathbf{U}=(1-\alpha)\mathbf{O}+\alpha\mathbf{P}$. (c) The EMA-teacher tracks the student's parameter trajectory with a temporal lag.}
  \label{fig:method}
  \vspace{-2mm}
\end{figure}

We first characterize the information retained by the EMA-teacher, derive the proposed update, analyze memory efficiency and sharded implementation, and present convergence guarantees. Throughout, we consider a student hidden weight matrix $\mathbf{W}_S\in\mathbb{R}^{A\times B}$ and its corresponding teacher matrix $\mathbf{W}_T$. Each matrix is processed independently. Figure~\ref{fig:method} illustrates the proposed framework.


\subsection{The Teacher Stores Applied Updates}
\label{sec:ema_trajectory}

Muon keeps a momentum buffer as large as each hidden matrix, while the EMA-teacher already stores a history of the same trajectory. If this history can take the role of momentum, the buffer becomes unnecessary. We therefore first characterize what the teacher stores and why it cannot simply replace momentum. Let $\mathbf{U}^{(s)}$ denote the applied update direction at step $s$, with $\mathbf{W}_S^{(s+1)}=\mathbf{W}_S^{(s)}-\eta c_{A,B}\mathbf{U}^{(s)}$, assuming constant learning rate $\eta$ and zero weight decay. Unrolling the EMA update in Eq.~\eqref{eq:ema_update}, initialized with $\mathbf{W}_T^{(0)}=\mathbf{W}_S^{(0)}$, gives (Lemma~\ref{lem:ema_history})
\begin{equation}
\small
\begin{aligned}
\mathbf{W}_T^{(s)}-\mathbf{W}_S^{(s)}
&=\tau\eta c_{A,B}
\sum_{j=0}^{s-1}\tau^j\mathbf{U}^{(s-1-j)}.
\end{aligned}
\label{eq:trajectory_history}
\end{equation}
Thus, the teacher lags behind the student, with its displacement encoding a discounted history of applied update directions. Unlike Muon's momentum, which accumulates raw gradients, this history contains optimizer outputs that have already been orthogonalized under standard Muon. Directly using this displacement as momentum would therefore mix historical updates with current gradients, potentially reinforcing stale directions as the self-distillation target evolves. Moreover, combining both signals before Newton--Schulz entangles their contributions through nonlinear orthogonalization, making their relative influence difficult to control. \our{} instead processes the two signals separately and combines them after normalization, explicitly controlling the historical contribution through a mixing weight.
\vspace{-2mm}
\subsection{The \our{} Update}
\label{sec:reuz_update}

\paragraph{Gradient branch.}
We eliminate Muon's momentum buffer and orthogonalize the current student gradient $\mathbf{G}=\nabla_{\mathbf{W}_S}\mathcal{L}(\theta_S)$:
\begin{equation}
\mathbf{O}=\mathrm{NS}_K(\mathbf{G}).
\label{eq:gradient_branch}
\end{equation}
This corresponds to Muon with $\mu=0$. Since the EMA-teacher and its distillation targets evolve during training, accumulated gradients may lag behind the current objective (Section~\ref{sec:exp_ablation}).
\vspace{-1.5mm}
\paragraph{Teacher branch.}
Let $\mathbf{D}=\mathbf{W}_T-\mathbf{W}_S$ denote the teacher--student displacement. We remove its component parallel to the student weights:
\begin{equation}
\small
\mathbf{H}=\mathbf{D}-
\frac{\langle\mathbf{D},\mathbf{W}_S\rangle_{\mathrm F}}
{\|\mathbf{W}_S\|_{\mathrm F}^{2}+\varepsilon}\mathbf{W}_S.
\label{eq:radial_projection}
\end{equation}
This suppresses radial changes associated with parameter scaling, including weight decay, while retaining the remaining trajectory information. We normalize the residual by its estimated largest singular value: \vspace{-3mm}
\begin{equation}
\small
\mathbf{P}=
\frac{\mathbf{H}}
{\max(\widehat{\sigma}_{\max}(\mathbf{H}),\varepsilon)}.
\label{eq:spectral_normalization}
\end{equation}
This brings its spectral norm approximately to one, comparable to the orthogonalized gradient $\mathbf{O}$. We estimate $\sigma_{\max}$ using $K_{\mathrm{PI}}$ power iterations, warm-started from the previous right singular vector, retaining only one vector of length $B$ per matrix.

\paragraph{Mixing after orthogonalization.}
We combine the orthogonalized gradient and normalized teacher directions:\vspace{-2mm}
\begin{align}
\mathbf{U}&=(1-\alpha)\mathbf{O}+\alpha\mathbf{P},
\label{eq:post_mix}\\
\mathbf{W}_S&\leftarrow(1-\eta\lambda)\mathbf{W}_S
-\eta c_{A,B}\mathbf{U}.
\label{eq:reuz_update_final}
\end{align}
Here, $\alpha\in[0,1]$ controls the teacher contribution, with $\|\alpha\mathbf{P}\|_2\leq\alpha$ under exact spectral normalization and an approximate bound with power iteration. Setting $\alpha=0$ recovers momentum-free Muon. The EMA-teacher is updated afterward using Eq.~\eqref{eq:ema_update}.

\begin{algorithm}[t]
\caption{One optimization step of \our{}}
\label{alg:reuz_muon}
\small
\begin{algorithmic}[1]
\Require Student $\theta_S$, EMA-teacher $\theta_T$, loss $\mathcal{L}$, mixing weight $\alpha$, EMA decay $\tau$, learning rate $\eta$, weight decay $\lambda$
\State Compute student gradients $\nabla_{\theta_S}\mathcal{L}$
\For{each Muon-optimized hidden matrix $\ell$}
    \State $\mathbf{G}_{\ell}\gets\nabla_{\mathbf{W}_{S,\ell}}\mathcal{L}$
    \State $\mathbf{O}_{\ell}\gets\mathrm{NS}_K(\mathbf{G}_{\ell})$
    \State $\mathbf{D}_{\ell}\gets\mathbf{W}_{T,\ell}-\mathbf{W}_{S,\ell}$
    \State Compute $\mathbf{H}_{\ell}$ using Eq.~\eqref{eq:radial_projection}
    \State Estimate $\widehat{\sigma}_{\max}(\mathbf{H}_{\ell})$ by power iteration
    \State Compute $\mathbf{P}_{\ell}$ using Eq.~\eqref{eq:spectral_normalization}
    \State $\mathbf{U}_{\ell}\gets(1-\alpha)\mathbf{O}_{\ell}+\alpha\mathbf{P}_{\ell}$
    \State Update $\mathbf{W}_{S,\ell}$ in place via Eq.~\eqref{eq:reuz_update_final}
    \State Release temporary matrix buffers
\EndFor
\State Update remaining student parameters with AdamW
\State $\theta_T\gets\tau\theta_T+(1-\tau)\theta_S$
\end{algorithmic}
\end{algorithm}





\subsection{Memory and Sharded Implementation}
\label{sec:memory_implementation}

\our{} reduces the persistent states of each Muon-optimized hidden matrix from $\{\mathbf{W}_S,\mathbf{W}_T,\mathbf{M}\}$ to $\{\mathbf{W}_S,\mathbf{W}_T\}$, eliminating the momentum buffer. Including gradient storage $\mathbf{G}$, this reduces the matrix-sized tensors from 4 to 3. Importantly, $\mathbf{O}$, $\mathbf{D}$, $\mathbf{H}$, $\mathbf{P}$, and $\mathbf{U}$ are temporary computational tensors, not additional parameters or persistent states. As summarized in Algorithm~\ref{alg:reuz_muon}, they are computed one matrix at a time, with buffers reused or released after updating $\mathbf{W}_S$ in place. Only a power-iteration vector of length $B$ per matrix is retained for warm-starting. Thus, the temporary workspace scales with the processed matrix size rather than requiring additional model-wide buffers. We report measured persistent memory in Section~\ref{sec:exp_efficiency} and detailed accounting in Appendix~\ref{app:projection}.

We use Fully Sharded Data Parallel (FSDP)~\citep{zhao2023fsdp}, with matching sharding layouts for student and EMA-teacher parameters. The displacement $\mathbf{D}=\mathbf{W}_T-\mathbf{W}_S$ is computed locally, while the Frobenius inner products for radial projection are all-reduced across ranks. Full matrices $\mathbf{G}$ and $\mathbf{H}$ are temporarily gathered for Newton--Schulz orthogonalization and power iteration, respectively. Each rank then updates only its corresponding shard of $\mathbf{W}_S$, followed by the sharded EMA update. With consistent global reductions and power-iteration initialization, the sharded implementation matches the unsharded computation up to numerical differences (Appendix~\ref{app:implementation}).

%% file: sections/3.1_theorem.tex
\subsection{Theoretical Analysis}
\label{sec:theoretical_analysis}

We now give the guarantees behind \our{}; Appendix~\ref{app:theory} contains the proofs and the full assumptions. All results hold for any differentiable host loss with a parameter-aligned EMA-teacher.

\begin{lemma}[EMA trajectory]
\label{lem:ema_history}
If $\mathbf{W}_S^{(s+1)}=\mathbf{W}_S^{(s)}-\eta c_{A,B}\mathbf{U}^{(s)}$, $\mathbf{W}_T^{(s+1)}=\tau\mathbf{W}_T^{(s)}+(1-\tau)\mathbf{W}_S^{(s+1)}$ and $\mathbf{W}_T^{(0)}=\mathbf{W}_S^{(0)}$, then Eq.~\eqref{eq:trajectory_history} holds for every $s\ge1$.
\end{lemma}

\begin{lemma}[Temporal averaging]
\label{lem:ema_variance}
Let $\overline{\mathbf{U}}^{(s)}=(1-\tau)\sum_{j<s}\tau^{j}\mathbf{U}^{(s-1-j)}$ be the rescaled displacement. If locally $\mathbf{U}^{(s)}=\mathbf{d}+\boldsymbol{\xi}^{(s)}$ with a fixed direction $\mathbf{d}$ and zero-mean, temporally uncorrelated noise of variance $\sigma_{\xi}^2$, then $\mathbb{E}[\overline{\mathbf{U}}^{(s)}]=(1-\tau^{s})\mathbf{d}$ and the variance of $\overline{\mathbf{U}}^{(s)}$ is $\frac{1-\tau}{1+\tau}(1-\tau^{2s})\sigma_{\xi}^2$.
\end{lemma}

Lemma~\ref{lem:ema_variance} shows that the teacher holds a smoothed estimate of persistent motion, with the noise variance reduced by a factor of up to $(1-\tau)/(1+\tau)$. The closer $\tau$ is to 1, the more stable the teacher direction is than any single step.

\begin{theorem}[Convergence of \our{}]
\label{thm:reuz_convergence}
Under A1--A4 with constant step $\eta$, $\lambda{=}0$ and $c_{A,B}$ absorbed into $\eta$, let $\gamma_R=a_R-\frac{L\eta}{2}b_R$. If $\gamma_R>0$ and $0<2\mu_{\mathrm{PL}}\eta\gamma_R<1$, then with $q_R=1-2\mu_{\mathrm{PL}}\eta\gamma_R$,
\begin{equation}
  \mathbb{E}[\Delta_{s+1}]\le q_R\,\mathbb{E}[\Delta_s]+\tfrac{L\eta^2}{2}\sigma_R^2+2\delta_s .
  \label{eq:main_recursion}
\end{equation}
Hence $\mathbb{E}[\Delta_s]$ converges linearly at rate $q_R$ to a neighborhood of size $O\big(\frac{L\eta\sigma_R^2}{\mu_{\mathrm{PL}}\gamma_R}+\frac{\bar\delta}{\mu_{\mathrm{PL}}\eta\gamma_R}\big)$ when $\delta_s\le\bar\delta$.
\end{theorem}
Proposition and Corollary in the Appendix~\ref{sec:app_theoretical_analysis} further shows where gain must come from and when the reuse of the teacher parameters helps.

%% file: sections/4_experiments.tex
\section{Experiments}
\label{sec:experiments}
We evaluate \our{} as a drop-in optimizer for two self-distillation hosts: fine-tuning with \base{} and on-policy post-training with \sdpo{}. Our evaluation covers task accuracy, retention of prior capabilities, convergence speed and memory. Finally, ablations isolate each design choice, and we test the theory of Section~\ref{sec:theoretical_analysis} against empirical measurements.

\input{sections/table1}

\subsection{Experimental Setup}
\label{sec:setup}
\noindent\textbf{Tasks and data.}
For fine-tuning with self-distillation, we follow \base{}~\citep{shenfeld2026sdft} on Qwen2.5-7B-Instruct~\citep{qwen2024qwen25} with its two tasks: tool use from ToolAlpaca~\citep{tang2023toolalpaca} and science questions from SciKnowEval~\citep{feng2024sciknoweval}. For on-policy post-training, we follow \sdpo{}~\citep{hubotter2026sdpo} on Olmo-3-7B-Instruct~\citep{olmo2025olmo3} with the biology and materials subsets of SciKnowEval. In both settings, we use the default splits and metrics of the original papers.

\noindent\textbf{Baselines.}
For \base{}, we compare with SFT, SFT with re-invoked prior knowledge, DFT~\citep{wu2025dft} and \base{} with AdamW, as reported by \citet{shenfeld2026sdft}. We additionally run \base{} with Muon and with MuonClip, which adds the query--key logit clipping (QK-Clip) of Kimi~K2~\citep{kimi2025k2}. Both Muon variant use $\mu{=}0$, their stronger setting; momentum $\mu{=}0.95$ is evaluated as an ablation (Table~\ref{tab:ablation}). For \sdpo{}, we compare with Group Relative Policy Optimization (GRPO;~\citealp{shao2024deepseekmath}), on-policy GRPO and \sdpo{} with AdamW, as reported by \citet{hubotter2026sdpo}, and run \sdpo{} with Muon, MuonClip and \our{}. The training and evaluation details are provide in the Appendix~\ref{sec:app_train_eval} and Appendix~\ref{app:hyperparameters} lists all hyperparameter settings.

\subsection{Fine-Tuning with Self-Distillation}
\label{sec:exp_finetuning}

\noindent\textbf{Task accuracy.}
Table~\ref{tab:continual_overall} shows that \our{} achieves the highest  new-task accuracy 
\input{sections/table_sdpo}
on both tasks: 74.2\% on tool use and 71.8\% on science, compared with 70.6\% and 70.2\% for \base{} with AdamW. It improves on momentum-free Muon by 5.1 and 2.8 points, respectively. The two methods share the same 32-bit teacher and differ only in whether the teacher branch is reused, so we attribute this gain to teacher reuse. \our{} also outperforms MuonClip without any logit clipping. Across both host models, \our{} ranks first among the Muon variants in all six task and budget settings (Tables~\ref{tab:continual_overall} and~\ref{tab:sdpo}).

\noindent\textbf{Prior capabilities.}
\our{} changes only the optimizer, so the question is whether its faster learning comes at the cost of prior capabilities. Retention is set mainly by the host loss: the prior-capability averages of all self-distillation runs lie within 2.4 points of one another, and those of \our{} are 11.9 and 6.7 points above SFT (Table~\ref{tab:continual_overall}). On science, \our{} has the best HumanEval score (70.7) and a higher average than \base{} with AdamW (65.3 against 64.5). On tool use, where \our{} gains the most on the new task, all Muon variants average 1.1--2.4 points below AdamW, mostly on IFEval. Even so \our{} retains 96\% of the AdamW average (63.0 against 65.4). \our{} thus improves new-task accuracy through optimization alone while preserving the retention advantage of self-distillation over SFT.

\begin{figure}[t]
  \centering
  \includegraphics[width=\linewidth]{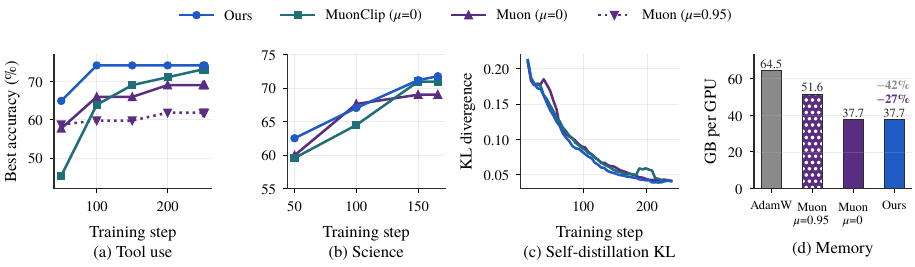}
  \vspace{-5mm}
  \caption{Training dynamics with \base{}. (a, b) Best accuracy reached up to each step. (c) Self-distillation KL at learning rate $5{\times}10^{-5}$. (d) Persistent memory per GPU between steps, with a bfloat16 mixed-precision student (32-bit master weights) and a 32-bit teacher for every optimizer. 
  }
  \label{fig:dynamics}
  \vspace{-2mm}
\end{figure}

\noindent\textbf{Convergence.}
Figure~\ref{fig:dynamics}(a)--(c) shows that \our{} learns faster (25-step averages for the KL). On tool use, it reaches 74.2\% after 100 of 252 steps, whereas MuonClip needs 250 steps to reach 73.2\% and Muon does not exceed 69.1\%. On science, it leads at 50 steps and ends highest. At the same learning rate, its self-distillation KL is lower than that of both baselines for most of training, as Corollary~\ref{cor:faster_than_baseline} predicts when the teacher branch raises the effective descent coefficient.
\vspace{-3mm}
\subsection{On-Policy Post-Training}
\label{sec:exp_onpolicy}

It is evident from Table~\ref{tab:sdpo}, \our{} is the most accurate optimizer for \sdpo{} on both datasets and budgets, 2.0 points above AdamW on average. Its advantage grows with training time: its margin over \sdpo{} with AdamW is 0.7–1.1 points after 1 hour and widens to 2.4–3.7 points after 5 hours. This trend is consistent with the gain arising from information accumulated over the optimization trajectory rather than from any single step update. On Materials, \our{} outperforms all baselines (82.8\% after 5 hours). On Biology it is second only to GRPO, which uses a different training signal.
\vspace{-2mm}
\subsection{Memory and Compute}
\label{sec:exp_efficiency}

\begin{figure}[t]
  \centering
  \includegraphics[width=\linewidth]{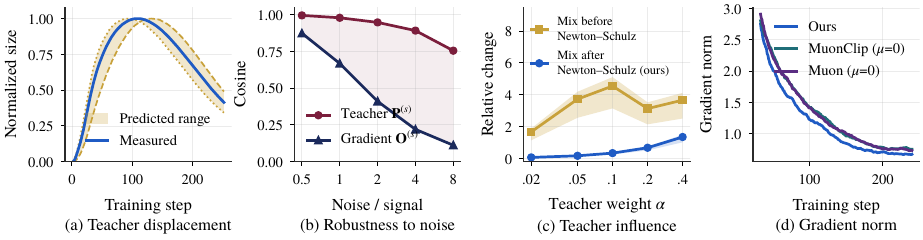}
  \vspace{-6mm}
  \caption{Analysis of the teacher branch. (a) Measured and predicted teacher--student displacement during training. (b) Cosine with the mean gradient direction under gradient noise on a Qwen2.5-7B-Instruct weight matrix. (c) Change of the update caused by the teacher direction. (d) Gradient norm during training.}
  \label{fig:analysis}
  \vspace{-2mm}
\end{figure}

\begin{wrapfigure}{r}{0.6\columnwidth}
  \centering
  \vspace{-4mm}
  \includegraphics[width=\linewidth]{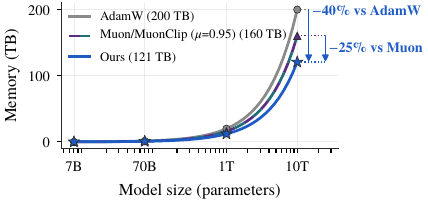}
  \vspace{-3mm}
  \caption{Projected persistent memory of self-distillation versus model size, with all states in 32 bits and gradients included.}
  \label{fig:projection}
\end{wrapfigure}
Figure~\ref{fig:dynamics}(d) reports persistent memory per GPU, measured between steps when gradients are freed, with every optimizer using a 32-bit EMA-teacher. Because \our{} keeps no optimizer state for hidden matrices, it needs only 37.7\,GB, 13.9\,GB (27\%) less than Muon with its standard momentum $\mu{=}0.95$ and 26.8\,GB (42\%) less than AdamW. Momentum-free Muon and MuonClip need the same 37.7\,GB but are 0.8--5.1 points less accurate (Table~\ref{tab:continual_overall}): \our{} therefore obtains the benefit of momentum without storing a momentum buffer. At 10T parameters, \our{} needs 121\,TB, 25\% less than Muon or MuonClip with momentum (160\,TB) and 40\% less than AdamW (200\,TB; Figure~\ref{fig:projection}, Appendix~\ref{app:projection}). The additional cost is one power iteration and one extra gather per matrix. The optimizer step takes 2.3\,s, about 2\% of a training step dominated by generation.

\subsection{Ablations and Analysis}
\label{sec:exp_ablation}

\input{sections/table_ablation}

We ablate each component of \our{} with \base{} on both tasks (Table~\ref{tab:ablation}): the optimizer state it replaces, the teacher weight $\alpha$, the construction of the teacher branch, and its combination with logit clipping. All ablations follow the setting of Table~\ref{tab:continual_overall}, with $\alpha{=}0.1$ unless varied, and every block includes the full method as reference. The Appendix~\ref{sec:app_exp_ablation} provides the detailed discussion about the ablation provide in the Table~\ref{tab:continual_overall}.

\vspace{-1.5mm}
\subsection{Empirical Checks of the Theory}
\label{sec:exp_theory}

Figure~\ref{fig:analysis} tests each theoretical statement with a matching measurement. The details are discussed in the Appendix~\ref{sec:app_exp_theory} where we empirically justify each theoretical claims.

%% file: sections/table1.tex
\begin{table}[t]
\centering

\small
\setlength{\tabcolsep}{7.2pt}
\renewcommand{\arraystretch}{1.08}
\resizebox{\linewidth}{!}{
\begin{tabular}{@{}cll|c|cccccc|c@{}}
\toprule

\multicolumn{1}{c}{\textbf{Task}} & \multicolumn{1}{c}{\textbf{Method}} 
& \multicolumn{1}{c}{\textbf{Optimizer}} &
\multicolumn{1}{c|}{\textbf{W/O CL}} &
\multicolumn{7}{c}{\textbf{CL Tasks}} \\
\cmidrule(lr){4-4}
\cmidrule(lr){5-11}
& & &
\textbf{Accuracy} & \textbf{HS} & \textbf{HE} &
\textbf{IFE} & \textbf{MMLU} & \textbf{TQA} &
\textbf{Wino} & \textbf{Avg.} \\
\midrule
\multirow{8}{*}{\rotatebox[origin=c]{90}{\textbf{Science}}}
& Base model\pub & -- & 32.1 & \best{62.0} & 65.8 & \best{74.3} & \best{71.7} & \best{47.9} & 71.1 & \best{65.5} \\
& SFT\pub & AdamW & 66.2 & 55.0 & 54.8 & 35.3 & 64.6 & 36.8 & \best{73.7} & 53.4 \\
& SFT + re-invoke\pub & AdamW & 66.0 & \second{61.6} & 63.4 & 52.9 & 68.7 & 45.2 & 70.0 & 60.3 \\
& DFT\pub & AdamW & 54.8 & 57.6 & 67.0 & 60.4 & 69.4 & 38.8 & 68.2 & 60.2 \\
& \base{}\pub & AdamW & 70.2 & 60.9 & \second{68.9} & 66.8 & 70.7 & 46.5 & \second{73.1} & 64.5 \\
& \base{} & Muon ($\mu{=}0$) & 69.0 & 61.5 & 68.3 & 67.9 & \second{71.6} & \second{47.7} & 71.9 & 64.8 \\
& \base{} & MuonClip ($\mu{=}0$) & \second{71.0} & \second{61.6} & 67.1 & \second{72.4} & \best{71.7} & 47.5 & 72.1 & \second{65.4} \\
& \textbf{\base{}} & \textbf{\our{} (ours)} & \best{71.8} & 61.2 & \best{70.7} & 69.3 & \second{71.6} & 46.5 & 72.6 & 65.3 \\
\midrule
\multirow{8}{*}{\rotatebox[origin=c]{90}{\textbf{Tool use}}}
& Base model\pub & -- & 42.9 & \best{62.0} & 65.8 & \best{74.3} & \best{71.7} & \second{47.9} & 71.1 & \best{65.5} \\
& SFT\pub & AdamW & 63.2 & 57.3 & 50.0 & 49.8 & 70.2 & 37.5 & \best{73.1} & 56.3 \\
& SFT + re-invoke\pub & AdamW & 63.1 & \second{61.7} & \best{68.9} & 59.1 & 71.5 & \best{49.1} & 71.6 & 63.7 \\
& DFT\pub & AdamW & 64.2 & 59.7 & 61.4 & 60.2 & \second{71.6} & 40.2 & 71.5 & 60.8 \\
& \base{}\pub & AdamW & 70.6 & 61.6 & \second{68.3} & \second{71.9} & 71.5 & 47.3 & 71.7 & \second{65.4} \\
& \base{} & Muon ($\mu{=}0$) & 69.1 & 61.4 & 65.2 & 69.3 & 71.5 & 47.0 & 71.5 & 64.3 \\
& \base{} & MuonClip ($\mu{=}0$) & \second{73.2} & 61.5 & 61.0 & 69.5 & \second{71.6} & 46.0 & 71.6 & 63.5 \\
& \textbf{\base{}} & \textbf{\our{} (ours)} & \best{74.2} & 61.1 & 63.4 & 63.7 & 71.4 & 46.1 & \second{72.3} & 63.0 \\
\bottomrule
\end{tabular}
}
\caption{
The accuracy without continual learning (W/O CL) and with continual learning (CL) task for SDFT (Qwen2.5-7B). HS, HE, IFE, TQA, and Wino denote HellaSwag, HumanEval, IFEval, TruthfulQA, and WinoGrande; Avg.\ is the mean across previous tasks. \pub{} denotes results from SDFT, with averages recomputed. \our{} shows significant improvement without CL setting, however for the CL setting its highly competitive.
}
\label{tab:continual_overall}
\vspace{-3mm}
\end{table}

%% file: sections/table_sdpo.tex
\begin{wraptable}{r}{0.5\textwidth}
\centering

\small
\setlength{\tabcolsep}{4pt}
\renewcommand{\arraystretch}{1.08}
\resizebox{\linewidth}{!}{%
\begin{tabular}{@{}cllcc@{}}
\toprule
& \textbf{Method} & \textbf{Optimizer} & \textbf{1\,h} & \textbf{5\,h} \\
\midrule
\multirow{7}{*}{\rotatebox[origin=c]{90}{\textbf{Biology}}}
& Base model\pub & -- & \multicolumn{2}{c}{16.2} \\
& GRPO\pub & AdamW & 35.6 & \best{55.8} \\
& GRPO (on-policy)\pub & AdamW & \best{49.8} & 49.8 \\
& \sdpo{}\pub & AdamW & 48.0 & 52.8 \\
& \sdpo{} & Muon ($\mu{=}0$) & 48.7 & 53.1 \\
& \sdpo{} & MuonClip ($\mu{=}0$) & 48.5 & 54.4 \\
& \textbf{\sdpo{}} & \textbf{\our{} (ours)} & \second{49.1} & \second{55.2} \\
\midrule
\multirow{7}{*}{\rotatebox[origin=c]{90}{\textbf{Materials}}}
& Base model\pub & -- & \multicolumn{2}{c}{36.7} \\
& GRPO\pub & AdamW & 70.9 & 75.0 \\
& GRPO (on-policy)\pub & AdamW & 73.3 & 73.5 \\
& \sdpo{}\pub & AdamW & 73.7 & 79.1 \\
& \sdpo{} & Muon ($\mu{=}0$) & 73.2 & 81.3 \\
& \sdpo{} & MuonClip ($\mu{=}0$) & \second{73.8} & \second{81.5} \\
& \textbf{\sdpo{}} & \textbf{\our{} (ours)} & \best{74.4} & \best{82.8} \\
\bottomrule
\end{tabular}}
\caption{On-policy post-training with SDPO (Olmo-3-7B). Best avg@16 accuracy (\%) within 1 and 5 hours. \pub{} from SDPO. Best in bold, second-best underlined.}
\label{tab:sdpo}
\end{wraptable}

%% file: sections/table_ablation.tex
\begin{wraptable}[17]{r}{0.5\textwidth}
\centering
\vspace{-10mm}

\small
\setlength{\tabcolsep}{5pt}
\resizebox{\linewidth}{!}{
\begin{tabular}{@{}l cc@{}}
\toprule
\textbf{Variant} & \textbf{Tool use} & \textbf{Science} \\
\midrule
\multicolumn{3}{@{}l}{\textit{(a) Optimizer state}} \\
Muon ($\mu{=}0.95$) & 66.9 & 68.3 \\
MuonClip ($\mu{=}0.95$) & 67.0 & 68.8 \\
Muon ($\mu{=}0$) & 69.1 & 69.0 \\
MuonClip ($\mu{=}0$) & 73.2 & 71.0 \\
\textbf{\our{} (ours)} & \textbf{74.2} & \textbf{71.8} \\
\midrule
\multicolumn{3}{@{}l}{\textit{(b) Teacher weight $\alpha$}} \\
$\alpha{=}0$ (Muon, $\mu{=}0$) & 69.1 & 69.0 \\
$\alpha{=}0.05$ & 67.0 & 66.5 \\
\textbf{$\boldsymbol{\alpha{=}0.1}$ (ours)} & \textbf{74.2} & \textbf{71.8} \\
$\alpha{=}0.2$ & 68.0 & 66.8 \\
$\alpha{=}0.4$ & 67.0 & 65.4 \\
\midrule
\multicolumn{3}{@{}l}{\textit{(c) Teacher branch}} \\
\textbf{\our{} (ours)} & \textbf{74.2} & \textbf{71.8} \\
Mix before \ns{} & 72.5 & 69.2 \\
Frobenius instead of spectral norm & 71.8 & 68.5 \\
Keep radial part & 70.5 & 67.3 \\
Cosine gate on $\alpha$ & 69.2 & 66.8 \\
\midrule
\multicolumn{3}{@{}l}{\textit{(d) Logit clipping}} \\
\textbf{\our{} (ours)} & \textbf{74.2} & \textbf{71.8} \\
\our{} + QK-Clip (student only) & 66.0 & 70.6 \\
\our{} + QK-Clip (student \& teacher) & 67.3 & 70.1 \\
\bottomrule
\end{tabular}}
\caption{Ablations with \base{} (accuracy, \%).}
\label{tab:ablation}
\vspace{-3mm}
\end{wraptable}

%% file: sections/2_related_work.tex
\section{Related Work}
\label{sec:related-work}

Training on the model's own samples against a teacher improves adaptation~\citep{agarwal2024gkd,ko2024distillm}, with further advances in self-distillation~\citep{xu2024skd,wen2023fdiv}. Such approaches can reduce forgetting relative to SFT~\citep{yang2024sdft,shenfeld2025razor,chen2025retaining}, which often compromises prior capabilities~\citep{biderman2024lora,zheng2025spurious,shi2025clsurvey}. In particular, \base{}~\citep{shenfeld2026sdft} and \sdpo{}~\citep{hubotter2026sdpo} use privileged context with an EMA-teacher. These methods redesign the objective but retain standard optimizers, keeping 2 model-sized histories without exploiting their interaction. We preserve their objectives and modify only the optimizer, improving accuracy with both hosts.

Muon~\citep{jordan2024muon} brings orthogonalized updates to LLMs~\citep{liu2025muon,kimi2025k2}, supported by analyses of matrix updates~\citep{amsel2025polar,large2024modular} and of its convergence~\citep{li2025muonnote,shen2025muonconv}. Other optimizers include SOAP~\citep{vyas2025soap}, Sophia~\citep{liu2024sophia} and Lion~\citep{chen2023lion}. Memory-efficient methods use low-rank projections~\citep{zhao2024galore,hao2024flora,zhu2025apollo}, quantized or shared states~\citep{li2023fourbit,dettmers2023qlora,zhang2025adammini}, block updates~\citep{luo2024badam,pan2024lisa}, or fused updates~\citep{lv2024lomo,malladi2023mezo}. Others introduce slow averages~\citep{pagliardini2025ademamix} or examine optimizer tuning~\citep{zhao2025deconstructing}. These approaches redesign the optimizer's \emph{own} state rather than reuse model copies already maintained during training. \our{} eliminates Muon's momentum state, saving 27\% of the measured memory at 7B and 25\% of the projected memory at 10T while improving accuracy.

EMA-teachers provide stable learning targets~\citep{tarvainen2017meanteacher,grill2020byol,caron2021dino}, extending to representation learning~\citep{oquab2024dinov2} and EMA scaling~\citep{busbridge2023ema,moralesbrotons2024ema}. Weight averaging also improves optimization and model selection~\citep{polyak1992averaging,izmailov2018swa,zhang2019lookahead,jang2024modelstock,rame2024warm,defazio2024schedulefree}, and momentum itself can be written as an average of iterates~\citep{defazio2020primal} or mixed with the current gradient~\citep{ma2019qhm}. These approaches treat the average as a model, a target or a dedicated buffer, without characterizing the information encoded by its displacement from current weights. We show that this gap stores applied updates, averages out noise, and provides a reusable optimized direction.

%% file: sections/5_conclusion.tex
\section{Conclusion}
\label{sec:conclusion}

We asked whether the EMA-teacher of self-distillation can replace Muon's momentum buffer. We showed that the teacher--student displacement is a discounted sum of applied updates (Lemma~\ref{lem:ema_history}) that averages out their noise (Lemma~\ref{lem:ema_variance}), while the orthogonalized gradient is already optimal for a single step (Proposition~\ref{prop:polar_alignment}); the teacher thus adds temporal information rather than a better instantaneous direction. On this basis, \our{} drops the momentum buffer and adds the radial-free, spectrally normalized displacement after orthogonalization with a bounded weight, with linear convergence under the PL condition. Empirically, it reached the highest accuracy among all optimizers with both \base{} and \sdpo{}, converged faster with \base{}, retained prior capabilities well above SFT and reduced memory. The ablations show that gradient momentum hurts when the target moves, whereas the teacher's history helps when it enters after orthogonalization and stays consistent with the student. Any method that keeps a parameter-aligned EMA model can lend it to the optimizer.

%% file: sections/6_limitations.tex
\section*{Limitations}
\label{sec:limitations}

Our experiments cover 7B-parameter models from 2 families and 4 tasks; larger models and further domains are left for future work. The convergence analysis relies on smoothness, PL and alignment assumptions that are standard for non-convex optimization but whose constants are not estimated for LLMs. \our{} applies when a parameter-aligned EMA-teacher already exists, as in \base{} and \sdpo{}; for methods without one, a teacher kept only to replace momentum would not save memory.

%% file: sections/7_ethics.tex
\section*{Ethics Statement}
\label{sec:ethics}

This work changes how a model is optimized, not what it is trained on, and it uses public datasets (ToolAlpaca, SciKnowEval) under their released licenses and splits. Models trained on science questions may still produce wrong answers about chemical, biological or material properties and should not be used for laboratory or safety decisions without expert checks. Reducing optimizer memory lowers the hardware needed for continual adaptation; we report our compute in Appendix~\ref{app:hyperparameters}. AI assistants were used only for language and writing refinement; the authors checked all content and remain responsible for it.

%% file: sections/appen_proof.tex
\subsection{Theoretical Analysis}
\label{sec:app_theoretical_analysis}

This section provides remaining theoretical analysis. The below proposition shows where this gain must come from.

\begin{proposition}[No instantaneous alignment gain]
\label{prop:polar_alignment}
Let $\mathbf{Q}$ be the polar factor of $\mathbf{G}\neq\mathbf{0}$. Then $\mathbf{Q}\in\arg\max_{\|\mathbf{X}\|_2\le1}\langle\mathbf{G},\mathbf{X}\rangle_{\mathrm F}$, with maximum $\|\mathbf{G}\|_{*}$. Hence any teacher direction with $\|\mathbf{P}\|_2\le1$ satisfies $\langle\mathbf{G},\mathbf{P}\rangle_{\mathrm F}\le\langle\mathbf{G},\mathbf{Q}\rangle_{\mathrm F}$.
\end{proposition}

So the \ns{} direction is already optimal for the current gradient, and the teacher branch contributes what a single gradient cannot: temporal information. To make this precise, write $F_s$ for the expected host loss at step $s$ (it moves with the teacher), $F_s^\star=\inf F_s$ and $\Delta_s=F_s(\mathbf{W}_S^{(s)})-F_s^\star$. We assume (A1) $L$-smoothness, (A2) the PL condition $\frac12\|\nabla F_s\|_{\mathrm F}^2\ge\mu_{\mathrm{PL}}(F_s-F_s^\star)$~\citep{karimi2016pl}, (A3) bounded teacher drift $\sup_{\mathbf{W}}|F_{s+1}-F_s|\le\delta_s$, and (A4) alignment $\mathbb{E}\langle\mathbf{g}^{(s)},\mathbf{U}^{(s)}\rangle\ge a_R\|\mathbf{g}^{(s)}\|^2$ with second moment $\mathbb{E}\|\mathbf{U}^{(s)}\|^2\le b_R\|\mathbf{g}^{(s)}\|^2+\sigma_R^2$, where $\mathbf{g}^{(s)}=\nabla F_s(\mathbf{W}_S^{(s)})$.

The below corollary further shows when reuse of the teacher parameters are faster:

\begin{corollary}[When reuse is faster]
\label{cor:faster_than_baseline}
Let a baseline optimizer (e.g.\ momentum-free Muon) satisfy A1--A4 with $\gamma_B>0$ at the same $\eta$. If $\gamma_R>\gamma_B$, the \our{} bound contracts strictly faster ($q_R<q_B$); if, in addition, its neighborhood term is smaller, it also reaches a lower error floor.
\end{corollary}

Corollary~\ref{cor:faster_than_baseline} shows that reuse helps when the teacher branch raises the effective descent coefficient, i.e.\ adds alignment with the true gradient over time without inflating the update's second moment.

\subsection{Theoretical Analysis: Proofs and Convergence}
\label{app:theory}

This appendix proves the results stated in Section~\ref{sec:theoretical_analysis} and
gives the full assumptions behind them. Notation follows
Section~\ref{sec:method}.

\subsubsection{Proof of Lemma~\ref{lem:ema_history}}
\label{app:proof_ema_history}

\begin{proof}
Define the displacement
\begin{equation}
    \mathbf{D}^{(s)}
    =
    \mathbf{W}_T^{(s)}-\mathbf{W}_S^{(s)}.
\end{equation}
Using the EMA update, Eq.~\eqref{eq:ema_update},
\begin{align}
    \mathbf{D}^{(s)}
    &=
    \tau\mathbf{W}_T^{(s-1)}
    +
    (1-\tau)\mathbf{W}_S^{(s)}
    -
    \mathbf{W}_S^{(s)}
    \nonumber\\
    &=
    \tau
    \left(
        \mathbf{W}_T^{(s-1)}-\mathbf{W}_S^{(s)}
    \right)
    \nonumber\\
    &=
    \tau
    \left(
        \mathbf{W}_T^{(s-1)}-\mathbf{W}_S^{(s-1)}
    \right)
    \nonumber\\
    &\quad+
    \tau
    \left(
        \mathbf{W}_S^{(s-1)}-\mathbf{W}_S^{(s)}
    \right)
    \nonumber\\
    &=
    \tau
    \left(
        \mathbf{D}^{(s-1)}
        +
        \eta c_{A,B}\,\mathbf{U}^{(s-1)}
    \right),
    \label{eq:ema_gap_recursion_proof}
\end{align}
where the last equality follows from
the student update of Lemma~\ref{lem:ema_history}. Since
$\mathbf{D}^{(0)}=\mathbf{0}$, recursively expanding
Eq.~\eqref{eq:ema_gap_recursion_proof} gives
\begin{equation}
    \mathbf{D}^{(s)}
    =
    \tau\eta c_{A,B}
    \sum_{j=0}^{s-1}
    \tau^j
    \mathbf{U}^{(s-1-j)},
\end{equation}
which is Eq.~\eqref{eq:trajectory_history}.
\end{proof}

\subsubsection{Proof of Lemma~\ref{lem:ema_variance}}
\label{app:proof_ema_variance}

\begin{proof}
Substituting the local model $\mathbf{U}^{(s)}=\mathbf{d}+\boldsymbol{\xi}^{(s)}$ of Lemma~\ref{lem:ema_variance} into
the definition of $\overline{\mathbf{U}}^{(s)}$,
\begin{align}
    \overline{\mathbf{U}}^{(s)}
    &=
    (1-\tau)
    \sum_{j=0}^{s-1}
    \tau^j
    \left(
        \mathbf{d}
        +
        \boldsymbol{\xi}^{(s-1-j)}
    \right)
    \nonumber\\
    &=
    (1-\tau)
    \left(
        \sum_{j=0}^{s-1}
        \tau^j
    \right)
    \mathbf{d}
    \nonumber\\
    &\quad+
    (1-\tau)
    \sum_{j=0}^{s-1}
    \tau^j
    \boldsymbol{\xi}^{(s-1-j)}.
\end{align}
Since
\begin{equation}
    (1-\tau)
    \sum_{j=0}^{s-1}
    \tau^j
    =
    1-\tau^s,
\end{equation}
taking expectations gives the stated mean. After subtracting the mean,
\begin{equation}
    \overline{\mathbf{U}}^{(s)}
    -
    \mathbb{E}[\overline{\mathbf{U}}^{(s)}]
    =
    (1-\tau)
    \sum_{j=0}^{s-1}
    \tau^j
    \boldsymbol{\xi}^{(s-1-j)}.
\end{equation}
Therefore,
\begin{align}
    &
    \mathbb{E}
    \left[
        \left\|
            \overline{\mathbf{U}}^{(s)}
            -
            \mathbb{E}[\overline{\mathbf{U}}^{(s)}]
        \right\|_{\mathrm F}^{2}
    \right]
    \nonumber\\
    &\quad=
    (1-\tau)^2
    \mathbb{E}
    \left[
        \left\|
            \sum_{j=0}^{s-1}
            \tau^j
            \boldsymbol{\xi}^{(s-1-j)}
        \right\|_{\mathrm F}^{2}
    \right].
\end{align}
All cross terms vanish by temporal uncorrelatedness, yielding
\begin{align}
    &
    \mathbb{E}
    \left[
        \left\|
            \overline{\mathbf{U}}^{(s)}
            -
            \mathbb{E}[\overline{\mathbf{U}}^{(s)}]
        \right\|_{\mathrm F}^{2}
    \right]
    \nonumber\\
    &\quad=
    (1-\tau)^2
    \sum_{j=0}^{s-1}
    \tau^{2j}\sigma_{\xi}^2
    \nonumber\\
    &\quad=
    (1-\tau)^2
    \frac{
        1-\tau^{2s}
    }{
        1-\tau^2
    }
    \sigma_{\xi}^2
    \nonumber\\
    &\quad=
    \frac{1-\tau}{1+\tau}
    \left(
        1-\tau^{2s}
    \right)
    \sigma_{\xi}^2,
\end{align}
which proves the variance statement of Lemma~\ref{lem:ema_variance}. Letting
$s\rightarrow\infty$ gives the stated limit.
\end{proof}

\begin{remark}
\label{rem:ema_variance_scope}
Lemma~\ref{lem:ema_variance} applies to the \emph{raw, rescaled} EMA
trajectory before the radial projection and spectral normalization in
\our{}. Since these operations are nonlinear, the factor
$(1-\tau)/(1+\tau)$ should not be interpreted as an exact variance reduction
for the final normalized direction $\mathbf{P}^{(s)}$. Moreover, for $\alpha>0$
the applied direction $\mathbf{U}^{(s)}$ itself contains $\mathbf{P}^{(s)}$, so its
noise is not exactly temporally uncorrelated; the lemma is exact for
$\alpha=0$ and a small-$\alpha$ approximation otherwise. The lemma instead
establishes that the state from which $\mathbf{P}^{(s)}$ is constructed already
contains a temporally smoothed estimate of persistent parameter motion.
\end{remark}

\subsubsection{Instantaneous alignment is not sufficient to explain the gain}
The following proposition is useful for interpreting the role of the EMA
branch.

\begin{restatement}[Proposition~\ref{prop:polar_alignment}]
Let $\mathbf{G}=\mathbf{L}_{G}\boldsymbol{\Sigma}\mathbf{V}_{G}^{\top}$ be a
compact singular value decomposition of $\mathbf{G}\neq\mathbf{0}$ and define
its exact polar factor
\begin{equation}
    \mathbf{Q}
    =
    \mathbf{L}_{G}\mathbf{V}_{G}^{\top}.
\end{equation}
Then
\begin{equation}
    \boxed{
    \mathbf{Q}
    \in
    \arg\max_{
        \|\mathbf{X}\|_2\leq 1
    }
    \left\langle
        \mathbf{G},
        \mathbf{X}
    \right\rangle_{\mathrm F},
    }
    \label{eq:polar_duality}
\end{equation}
and
\begin{equation}
    \left\langle
        \mathbf{G},
        \mathbf{Q}
    \right\rangle_{\mathrm F}
    =
    \|\mathbf{G}\|_*,
    \label{eq:polar_nuclear}
\end{equation}
where $\|\cdot\|_*$ denotes the nuclear norm. Therefore, for any
spectrally normalized teacher direction $\mathbf{P}$ satisfying
$\|\mathbf{P}\|_2\leq 1$,
\begin{equation}
    \left\langle
        \mathbf{G},
        \mathbf{P}
    \right\rangle_{\mathrm F}
    \leq
    \left\langle
        \mathbf{G},
        \mathbf{Q}
    \right\rangle_{\mathrm F}.
    \label{eq:teacher_vs_polar_alignment}
\end{equation}
\end{restatement}

\begin{proof}
By the duality between the nuclear and spectral norms,
\begin{equation}
    \|\mathbf{G}\|_*
    =
    \max_{\|\mathbf{X}\|_2\leq 1}
    \left\langle
        \mathbf{G},
        \mathbf{X}
    \right\rangle_{\mathrm F}.
\end{equation}
For $\mathbf{Q}=\mathbf{L}_{G}\mathbf{V}_{G}^{\top}$,
$\|\mathbf{Q}\|_2=1$, and
\begin{align}
    \left\langle
        \mathbf{G},
        \mathbf{Q}
    \right\rangle_{\mathrm F}
    &=
    \operatorname{tr}
    \left(
        \mathbf{G}^{\top}\mathbf{Q}
    \right)
    \nonumber\\
    &=
    \operatorname{tr}
    \left(
        \mathbf{V}_{G}
        \boldsymbol{\Sigma}
        \mathbf{L}_{G}^{\top}
        \mathbf{L}_{G}
        \mathbf{V}_{G}^{\top}
    \right)
    \nonumber\\
    &=
    \operatorname{tr}
    \left(
        \boldsymbol{\Sigma}
    \right)
    =
    \|\mathbf{G}\|_*.
\end{align}
Thus $\mathbf{Q}$ attains the spectral-norm-constrained maximum, proving
Eqs.~\eqref{eq:polar_duality}--\eqref{eq:polar_nuclear}. Since
$\|\mathbf{P}\|_2\leq 1$, Eq.~\eqref{eq:teacher_vs_polar_alignment} follows
immediately.
\end{proof}

\begin{remark}
Proposition~\ref{prop:polar_alignment} rules out an overly strong
interpretation of \our{}. In the idealized deterministic setting with an exact
polar factor, the teacher branch cannot improve the current gradient's
\emph{linear} alignment beyond the polar direction while remaining inside the
same spectral-norm ball. Any optimization advantage must therefore arise from
temporal information, stochastic denoising, finite Newton--Schulz
approximation, curvature-dependent second-order effects, or a combination of
these factors.
\end{remark}

\subsubsection{Finite-time convergence of \our{}}
We now analyze the complete mixed direction of Eq.~\eqref{eq:post_mix},
written $\mathbf{U}_R^{(s)}$ to distinguish it from baseline directions,
\begin{equation}
    \mathbf{U}_R^{(s)}
    =
    (1-\alpha)\mathbf{O}^{(s)}
    +
    \alpha\mathbf{P}^{(s)},
    \qquad
    \alpha\in[0,1),
    \label{eq:reuz_theory_direction}
\end{equation}
where $\mathbf{O}^{(s)}$ is the momentum-free Muon branch and
$\mathbf{P}^{(s)}$ is the normalized EMA-trajectory branch.

The host objective may itself move because the teacher changes. With
$\mathbf{W}$ denoting the Muon-eligible matrix under analysis, we therefore
write the expected host objective as
\begin{equation}
    F_s(\mathbf{W})
    =
    \mathbb{E}_{\mathcal{B}}
    \left[
    \mathcal{L}
    \left(
        \mathbf{W},
        \mathbf{W}_T^{(s)};
        \mathcal{B}
    \right)
    \right],
\end{equation}
with
\begin{equation}
    F_s^\star
    =
    \inf_{\mathbf{W}}F_s(\mathbf{W}),
    \qquad
    \Delta_s
    =
    F_s(\mathbf{W}_S^{(s)})-F_s^\star.
\end{equation}

We use the following standard local conditions over the region visited during
training.

\begin{enumerate}
    \item[\textbf{A1.}]
    \textbf{Smoothness.}
    Every $F_s$ is $L$-smooth with respect to the Frobenius norm:
    \begin{equation}
    \begin{split}
        F_s(\mathbf{Y})
        \leq{}&
        F_s(\mathbf{X})
        +
        \left\langle
            \nabla F_s(\mathbf{X}),
            \mathbf{Y}-\mathbf{X}
        \right\rangle_{\mathrm F}
        \\
        &+
        \frac{L}{2}
        \|\mathbf{Y}-\mathbf{X}\|_{\mathrm F}^{2}.
    \end{split}
        \label{eq:assumption_smoothness}
    \end{equation}

    \item[\textbf{A2.}]
    \textbf{PL condition.}
    There exists $\mu_{\mathrm{PL}}>0$ such that
    \begin{equation}
        \frac{1}{2}
        \left\|
            \nabla F_s(\mathbf{W})
        \right\|_{\mathrm F}^{2}
        \geq
        \mu_{\mathrm{PL}}
        \left(
            F_s(\mathbf{W})-F_s^\star
        \right)
        \label{eq:assumption_pl}
    \end{equation}
    for all iterates under consideration.

    \item[\textbf{A3.}]
    \textbf{Controlled teacher-induced objective drift.}
    There exists $\delta_s\geq 0$ such that
    \begin{equation}
        \sup_{\mathbf{W}}
        \left|
            F_{s+1}(\mathbf{W})
            -
            F_s(\mathbf{W})
        \right|
        \leq
        \delta_s.
        \label{eq:assumption_drift}
    \end{equation}

    \item[\textbf{A4.}]
    \textbf{Descent alignment and second moment.}
    Let
    \begin{equation}
        \mathbf{g}^{(s)}
        =
        \nabla F_s(\mathbf{W}_S^{(s)}).
    \end{equation}
    Conditioned on the optimization history $\mathcal{F}_s$, the \our{}
    direction satisfies
    \begin{equation}
        \mathbb{E}
        \left[
            \left\langle
                \mathbf{g}^{(s)},
                \mathbf{U}_R^{(s)}
            \right\rangle_{\mathrm F}
            \,\middle|\,
            \mathcal{F}_s
        \right]
        \geq
        a_R
        \|\mathbf{g}^{(s)}\|_{\mathrm F}^{2},
        \label{eq:assumption_alignment}
    \end{equation}
    and
    \begin{equation}
        \mathbb{E}
        \left[
            \|\mathbf{U}_R^{(s)}\|_{\mathrm F}^{2}
            \,\middle|\,
            \mathcal{F}_s
        \right]
        \leq
        b_R
        \|\mathbf{g}^{(s)}\|_{\mathrm F}^{2}
        +
        \sigma_R^2.
        \label{eq:assumption_second_moment}
    \end{equation}
\end{enumerate}

The constants in A4 have a direct interpretation. $a_R$ measures expected
alignment with the current true descent direction, $b_R$ measures how much
update energy grows with the gradient, and $\sigma_R^2$ collects residual
update energy that does not scale with the gradient. Because $\mathbf{O}^{(s)}$
and $\mathbf{P}^{(s)}$ are normalized, this residual includes their
gradient-independent magnitude as well as stochastic noise, and $a_R$ is
understood over a region in which $\|\mathbf{g}^{(s)}\|_{\mathrm F}$ is bounded.
A conservative set of valid coefficients can be
obtained from separate branch bounds. In particular, if
\begin{equation}
    \mathbb{E}
    \left[
        \langle
            \mathbf{g}^{(s)},
            \mathbf{O}^{(s)}
        \rangle_{\mathrm F}
        \mid\mathcal{F}_s
    \right]
    \geq
    a_O\|\mathbf{g}^{(s)}\|_{\mathrm F}^{2},
\end{equation}
\begin{equation}
    \mathbb{E}
    \left[
        \langle
            \mathbf{g}^{(s)},
            \mathbf{P}^{(s)}
        \rangle_{\mathrm F}
        \mid\mathcal{F}_s
    \right]
    \geq
    a_P\|\mathbf{g}^{(s)}\|_{\mathrm F}^{2},
\end{equation}
and
\begin{align}
    \mathbb{E}
    \left[
        \|\mathbf{O}^{(s)}\|_{\mathrm F}^{2}
        \mid\mathcal{F}_s
    \right]
    &\leq
    b_O\|\mathbf{g}^{(s)}\|_{\mathrm F}^{2}
    +
    \sigma_O^2,
    \\
    \mathbb{E}
    \left[
        \|\mathbf{P}^{(s)}\|_{\mathrm F}^{2}
        \mid\mathcal{F}_s
    \right]
    &\leq
    b_P\|\mathbf{g}^{(s)}\|_{\mathrm F}^{2}
    +
    \sigma_P^2,
\end{align}
where $a_P$ may be negative, then linearity of the inner product and
convexity of the squared norm imply
\begin{equation}
    a_R
    =
    (1-\alpha)a_O+\alpha a_P,
    \label{eq:a_reuz}
\end{equation}
\begin{equation}
\begin{split}
    b_R
    &=
    (1-\alpha)b_O+\alpha b_P,
    \\
    \sigma_R^2
    &=
    (1-\alpha)\sigma_O^2+\alpha\sigma_P^2.
\end{split}
    \label{eq:b_sigma_reuz}
\end{equation}
These are conservative because they do not exploit any favorable cancellation
between $\mathbf{O}^{(s)}$ and $\mathbf{P}^{(s)}$.

\begin{restatement}[Theorem~\ref{thm:reuz_convergence}]
Suppose A1--A4 hold and the \our{} student update is
\begin{equation}
    \mathbf{W}_S^{(s+1)}
    =
    \mathbf{W}_S^{(s)}
    -
    \eta\mathbf{U}_R^{(s)}
    \label{eq:theory_reuz_update}
\end{equation}
with constant $\eta>0$, i.e., Eq.~\eqref{eq:reuz_update_final} with
$\lambda=0$ and $c_{A,B}$ absorbed into $\eta$. Define the effective descent
coefficient
\begin{equation}
    \boxed{
    \gamma_R
    =
    a_R
    -
    \frac{L\eta}{2}b_R.
    }
    \label{eq:gamma_reuz}
\end{equation}
If
\begin{equation}
    \gamma_R>0
    \qquad\text{and}\qquad
    0<
    2\mu_{\mathrm{PL}}\eta\gamma_R
    <1,
    \label{eq:reuz_stepsize_condition}
\end{equation}
then
\begin{equation}
    \boxed{
    \mathbb{E}[\Delta_{s+1}]
    \leq
    q_R\,
    \mathbb{E}[\Delta_s]
    +
    \frac{L\eta^2}{2}\sigma_R^2
    +
    2\delta_s,
    }
    \label{eq:reuz_one_step_recursion}
\end{equation}
where
\begin{equation}
    q_R
    =
    1-
    2\mu_{\mathrm{PL}}\eta\gamma_R
    \in(0,1).
    \label{eq:q_reuz}
\end{equation}
Consequently, for any $n\geq 1$,
\begin{equation}
    \boxed{
    \begin{aligned}
    \mathbb{E}[\Delta_n]
    \leq{}&
    q_R^n\Delta_0
    \\
    &+
    \sum_{s=0}^{n-1}
    q_R^{n-1-s}
    \left(
        \frac{L\eta^2}{2}\sigma_R^2
        +
        2\delta_s
    \right).
    \end{aligned}
    }
    \label{eq:reuz_finite_time_general}
\end{equation}
If the objective drift is uniformly bounded by
$\delta_s\leq\bar{\delta}$, then
\begin{equation}
    \boxed{
    \begin{aligned}
    \mathbb{E}[\Delta_n]
    \leq{}&
    q_R^n\Delta_0
    \\
    &+
    \frac{
        \frac{L\eta^2}{2}\sigma_R^2
        +
        2\bar{\delta}
    }{
        1-q_R
    }
    \left(
        1-q_R^n
    \right).
    \end{aligned}
    }
    \label{eq:reuz_finite_time_uniform}
\end{equation}
In particular,
\begin{equation}
    \boxed{
    \limsup_{n\rightarrow\infty}
    \mathbb{E}[\Delta_n]
    \leq
    \frac{L\eta\sigma_R^2}
         {4\mu_{\mathrm{PL}}\gamma_R}
    +
    \frac{\bar{\delta}}
         {\mu_{\mathrm{PL}}\eta\gamma_R}.
    }
    \label{eq:reuz_error_floor}
\end{equation}
For a fixed objective, $\bar{\delta}=0$.
\end{restatement}

\begin{proof}
By $L$-smoothness of $F_s$ and
Eq.~\eqref{eq:theory_reuz_update},
\begin{align}
    F_s(\mathbf{W}_S^{(s+1)})
    \leq{}&
    F_s(\mathbf{W}_S^{(s)})
    -
    \eta
    \left\langle
        \mathbf{g}^{(s)},
        \mathbf{U}_R^{(s)}
    \right\rangle_{\mathrm F}
    \nonumber\\
    &+
    \frac{L\eta^2}{2}
    \|\mathbf{U}_R^{(s)}\|_{\mathrm F}^{2}.
    \label{eq:smoothness_descent_step}
\end{align}
Conditioning on $\mathcal{F}_s$ and applying
Eqs.~\eqref{eq:assumption_alignment} and
\eqref{eq:assumption_second_moment},
\begin{align}
    &\mathbb{E}
    \left[
        F_s(\mathbf{W}_S^{(s+1)})
        \mid\mathcal{F}_s
    \right]
    \leq
    \nonumber\\
    &\quad
    F_s(\mathbf{W}_S^{(s)})
    -
    \eta a_R
    \|\mathbf{g}^{(s)}\|_{\mathrm F}^{2}
    \nonumber\\
    &\quad+
    \frac{L\eta^2}{2}
    \left(
        b_R
        \|\mathbf{g}^{(s)}\|_{\mathrm F}^{2}
        +
        \sigma_R^2
    \right)
    \nonumber\\
    &=
    F_s(\mathbf{W}_S^{(s)})
    -
    \eta
    \left(
        a_R
        -
        \frac{L\eta}{2}b_R
    \right)
    \|\mathbf{g}^{(s)}\|_{\mathrm F}^{2}
    \nonumber\\
    &\quad+
    \frac{L\eta^2}{2}\sigma_R^2
    \nonumber\\
    &=
    F_s(\mathbf{W}_S^{(s)})
    -
    \eta\gamma_R
    \|\mathbf{g}^{(s)}\|_{\mathrm F}^{2}
    +
    \frac{L\eta^2}{2}\sigma_R^2.
    \label{eq:descent_gamma}
\end{align}
Using the PL inequality in Eq.~\eqref{eq:assumption_pl},
\begin{equation}
    \|\mathbf{g}^{(s)}\|_{\mathrm F}^{2}
    \geq
    2\mu_{\mathrm{PL}}
    \left(
        F_s(\mathbf{W}_S^{(s)})-F_s^\star
    \right)
    =
    2\mu_{\mathrm{PL}}\Delta_s.
\end{equation}
Substituting this into Eq.~\eqref{eq:descent_gamma} gives
\begin{equation}
\begin{split}
    &\mathbb{E}
    \left[
        F_s(\mathbf{W}_S^{(s+1)})-F_s^\star
        \mid\mathcal{F}_s
    \right]
    \\
    &\quad\leq
    q_R\Delta_s
    +
    \frac{L\eta^2}{2}\sigma_R^2.
\end{split}
    \label{eq:same_objective_recursion}
\end{equation}

It remains to account for the fact that the EMA-teacher changes the host
objective between steps. From A3,
\begin{equation}
    F_{s+1}(\mathbf{W}_S^{(s+1)})
    \leq
    F_s(\mathbf{W}_S^{(s+1)})
    +
    \delta_s.
    \label{eq:drift_upper_function}
\end{equation}
Moreover,
\begin{equation}
\begin{split}
    F_{s+1}^\star
    &=
    \inf_{\mathbf{W}}F_{s+1}(\mathbf{W})
    \\
    &\geq
    \inf_{\mathbf{W}}
    \left(
        F_s(\mathbf{W})-\delta_s
    \right)
    =
    F_s^\star-\delta_s.
\end{split}
    \label{eq:drift_lower_optimum}
\end{equation}
Combining
Eqs.~\eqref{eq:drift_upper_function} and
\eqref{eq:drift_lower_optimum},
\begin{align}
    \Delta_{s+1}
    &=
    F_{s+1}(\mathbf{W}_S^{(s+1)})
    -
    F_{s+1}^\star
    \nonumber\\
    &\leq
    F_s(\mathbf{W}_S^{(s+1)})
    -
    F_s^\star
    +
    2\delta_s.
    \label{eq:drift_gap_relation}
\end{align}
Taking total expectation in
Eq.~\eqref{eq:same_objective_recursion} and applying
Eq.~\eqref{eq:drift_gap_relation} yields
Eq.~\eqref{eq:reuz_one_step_recursion}.

Repeatedly applying the recursion gives
Eq.~\eqref{eq:reuz_finite_time_general}. When
$\delta_s\leq\bar{\delta}$,
the geometric-series identity
\begin{equation}
    \sum_{s=0}^{n-1}
    q_R^{n-1-s}
    =
    \frac{1-q_R^n}{1-q_R}
\end{equation}
gives Eq.~\eqref{eq:reuz_finite_time_uniform}. Finally,
using
\begin{equation}
    1-q_R
    =
    2\mu_{\mathrm{PL}}\eta\gamma_R
\end{equation}
and taking $n\rightarrow\infty$ yields
Eq.~\eqref{eq:reuz_error_floor}.
\end{proof}

\subsubsection{Comparison with a baseline optimizer}
The previous result immediately gives a sufficient condition for a faster
convergence guarantee than AdamW, momentum Muon, or momentum-free Muon,
provided the
corresponding baseline satisfies the same local regularity assumptions.

\begin{restatement}[Corollary~\ref{cor:faster_than_baseline}]
Let a baseline optimizer $B$ satisfy A1--A4 at the same step size $\eta$ with
coefficients
$a_B,b_B,\sigma_B^2$ and objective-drift bound
$\bar{\delta}_B$. Define
\begin{equation}
\begin{split}
    \gamma_B
    &=
    a_B
    -
    \frac{L\eta}{2}b_B,
    \\
    q_B
    &=
    1-
    2\mu_{\mathrm{PL}}\eta\gamma_B.
\end{split}
\end{equation}
If
\begin{equation}
    \boxed{
    \gamma_R>\gamma_B>0,
    }
    \label{eq:faster_condition}
\end{equation}
then
\begin{equation}
    q_R<q_B,
\end{equation}
so the \our{} bound contracts strictly faster per optimization step.

Furthermore, define the asymptotic upper-bound terms
\begin{equation}
    \Phi_R
    =
    \frac{L\eta\sigma_R^2}
         {4\mu_{\mathrm{PL}}\gamma_R}
    +
    \frac{\bar{\delta}_R}
         {\mu_{\mathrm{PL}}\eta\gamma_R},
    \label{eq:phi_reuz}
\end{equation}
and
\begin{equation}
    \Phi_B
    =
    \frac{L\eta\sigma_B^2}
         {4\mu_{\mathrm{PL}}\gamma_B}
    +
    \frac{\bar{\delta}_B}
         {\mu_{\mathrm{PL}}\eta\gamma_B}.
    \label{eq:phi_baseline}
\end{equation}
If, in addition,
\begin{equation}
    \boxed{
    \Phi_R<\Phi_B,
    }
    \label{eq:lower_floor_condition}
\end{equation}
then the \our{} convergence bound has both a faster transient contraction and a
strictly smaller asymptotic optimization neighborhood.
\end{restatement}

\begin{proof}
Equation~\eqref{eq:faster_condition} implies
\begin{equation}
    2\mu_{\mathrm{PL}}\eta\gamma_R
    >
    2\mu_{\mathrm{PL}}\eta\gamma_B,
\end{equation}
and therefore
\begin{equation}
\begin{split}
    q_R
    &=
    1-
    2\mu_{\mathrm{PL}}\eta\gamma_R
    \\
    &<
    1-
    2\mu_{\mathrm{PL}}\eta\gamma_B
    =
    q_B.
\end{split}
\end{equation}
Hence the geometric transient $q_R^n$ decays more rapidly than $q_B^n$.
The second statement follows directly from
Theorem~\ref{thm:reuz_convergence}: $\Phi_R$ and $\Phi_B$ are the respective
asymptotic upper bounds obtained by taking the limit superior of each
finite-time recursion. If $\Phi_R<\Phi_B$, the \our{} bound approaches a smaller
optimization neighborhood.
\end{proof}

For the fixed-objective case
$\bar{\delta}_R=\bar{\delta}_B=0$,
Eq.~\eqref{eq:lower_floor_condition} reduces to
\begin{equation}
    \boxed{
    \frac{\sigma_R^2}{\gamma_R}
    <
    \frac{\sigma_B^2}{\gamma_B}.
    }
    \label{eq:noise_descent_ratio}
\end{equation}
Thus the relevant quantity is not variance alone, but the ratio between
residual update energy and effective descent. This formalizes the
optimization mechanism suggested by Lemma~\ref{lem:ema_variance}: a temporally
smoothed trajectory is useful when the reduction in update noise is large
enough relative to any loss of instantaneous descent alignment.

\subsubsection{Iteration complexity}
Because the branches are normalized, $\sigma_R^2>0$ and the bound converges
to the neighborhood $\Phi_R$ rather than to zero. From
Eq.~\eqref{eq:reuz_finite_time_uniform},
\begin{equation}
    \mathbb{E}[\Delta_n]
    \leq
    q_R^n\Delta_0
    +
    \Phi_R.
\end{equation}
To reach $\mathbb{E}[\Delta_n]\leq\Phi_R+\epsilon_{\mathrm{tol}}$, it is
sufficient that
\begin{equation}
    n_R(\epsilon_{\mathrm{tol}})
    \geq
    \frac{
        \log(\Delta_0/\epsilon_{\mathrm{tol}})
    }{
        -\log q_R
    }.
    \label{eq:exact_iteration_complexity}
\end{equation}
For small $2\mu_{\mathrm{PL}}\eta\gamma_R$,
$-\log q_R\approx2\mu_{\mathrm{PL}}\eta\gamma_R$, giving
\begin{equation}
    n_R(\epsilon_{\mathrm{tol}})
    \approx
    \frac{1}{
        2\mu_{\mathrm{PL}}\eta\gamma_R
    }
    \log
    \frac{\Delta_0}{\epsilon_{\mathrm{tol}}}.
    \label{eq:approx_iteration_complexity}
\end{equation}
Accordingly, the local iteration-speedup predicted relative to a baseline,
each measured to within $\epsilon_{\mathrm{tol}}$ of its own neighborhood, is
approximately
\begin{equation}
    \boxed{
    \frac{
        n_B(\epsilon_{\mathrm{tol}})
    }{
        n_R(\epsilon_{\mathrm{tol}})
    }
    \approx
    \frac{
        \gamma_R
    }{
        \gamma_B
    }.
    }
    \label{eq:iteration_speedup}
\end{equation}
Equation~\eqref{eq:iteration_speedup} concerns optimization steps rather than
wall-clock time; the latter additionally depends on the per-step cost of each
optimizer.

\begin{remark}[What the theorem does and does not establish]
\label{rem:better_optimum}
Corollary~\ref{cor:faster_than_baseline} supports two claims under explicit
conditions: faster finite-time contraction and a lower optimization-error
bound. It does \emph{not} claim that \our{} universally reaches a better local
minimum in an arbitrary non-convex neural-network landscape. Higher downstream
accuracy can be consistent with the lower error bound and the temporal
smoothing of the raw trajectory discussed above, but the mapping from
optimization loss
to task accuracy is data- and objective-dependent and is therefore verified
empirically rather than assumed by the theorem.
\end{remark}

%% file: sections/appendix.tex
\subsection{Teacher Precision}
\label{app:fp32}

\our{} reads the teacher--student displacement $\mathbf{W}_T^{(s)}-\mathbf{W}_S^{(s)}$, so the teacher must actually follow the EMA of Eq.~\eqref{eq:ema_update}. With $\tau{=}0.99$, each step moves the teacher by $1\%$ of the displacement, and the displacement is only $0.1$--$0.3\%$ of the weights (Appendix~\ref{app:diagnostics}). The per-step change is therefore about $10^{-5}$ of a weight's magnitude, while bfloat16 can only represent relative changes larger than about $4\times10^{-3}$ (8 significand bits). In 16-bit storage most EMA steps round to zero. We measured this on a Qwen2.5-7B-Instruct key projection matrix with 250 momentum-free Muon steps (Figure~\ref{fig:appendix}(a); setup in Appendix~\ref{app:diagnostics}): the student moved by 2.15 in Frobenius norm, a 32-bit teacher by 1.37, and a bfloat16 teacher by only 0.66, with 79\% of its entries never changing. A 16-bit displacement would thus reflect rounding rather than the averaged trajectory that Lemmas~\ref{lem:ema_history} and~\ref{lem:ema_variance} describe. For the same reason, optimizer states such as Muon's momentum and AdamW's moments are usually kept in 32 bits~\citep{liu2025muon,li2023fourbit}; since the teacher takes over the role of optimizer state, we store it in 32 bits as well, and we do so for every optimizer we train in Tables~\ref{tab:continual_overall}--\ref{tab:ablation}, so that comparisons share the same teacher. The teacher's forward pass still runs in bfloat16, so the self-distillation loss is computed as in \base{}. The 32-bit teacher adds 2 bytes per parameter (7.6\,GB per GPU with 2 GPUs); the memory in Figure~\ref{fig:dynamics}(d) includes this cost.

\noindent\textbf{Precision setting.} Our precision differs from the original \base{} setting, which keeps the student and the teacher in bfloat16. To make sure that the gains of the Muon variants come neither from precision nor from run-to-run randomness, we train every optimizer in the same setting: a bfloat16 mixed-precision student with 32-bit master weights and a teacher stored in 32 bits, with forward passes in bfloat16. In this setting, \base{} and \sdpo{} with AdamW reproduce their reported accuracy within run-to-run variation, so the AdamW rows taken from \citet{shenfeld2026sdft} and \citet{hubotter2026sdpo} are comparable with our runs. All memory values in the paper are for a 32-bit teacher (Appendix~\ref{app:projection}).

\subsection{Implementation Details}
\label{app:implementation}

We build on the released \base{} code (TRL with colocated vLLM generation) and replace the optimizer. Hidden matrices are all two-dimensional weights inside the transformer blocks; embeddings, the output head, normalization weights and query/key/value biases use AdamW with the same learning rate, following \citet{liu2025muon}. Student and teacher are sharded with FSDP2 using the same wrapping policy, so each teacher shard matches a student shard. For every hidden matrix, the inner products in Eq.~\eqref{eq:radial_projection} and the norms are reduced across shards; the gradient (in bfloat16) and $\mathbf{H}^{(s)}$ are gathered, \ns{} and power iteration run on the full matrix on every rank, and each rank keeps its slice of $\mathbf{U}^{(s)}$. The power-iteration vector of each matrix is stored across steps (1 vector of length $B$, 4.5\,MB in total for Qwen2.5-7B); starting from a random vector, 8 iterations underestimate $\sigma_{\max}$ by about 8\%, and the warm start removes this error within 3 steps. We verified on a small model with 2 GPUs that (i) $\alpha{=}0$ is bit-identical to Muon with $\mu{=}0$, (ii) sharded and unsharded updates are identical, and (iii) for a single matrix the update matches an exact singular-value reference to $5\times10^{-7}$, with $\|\mathbf{P}^{(s)}\|_2{=}1$.

\subsection{Hyperparameters and Compute}
\label{app:hyperparameters}

\begin{table}[h]
\centering
\small
\setlength{\tabcolsep}{4pt}
\begin{tabular}{@{}ll@{}}
\toprule
\textbf{Setting} & \textbf{Value} \\
\midrule
Model & Qwen2.5-7B-Instruct \\
Hardware & 2$\times$H100 80\,GB, FSDP2 \\
Prompts per step & 32 (1 rollout each) \\
Epochs (steps) & 2 (252 tool use, 166 science) \\
Learning rate & $\{3,5\}{\times}10^{-5}$, cosine \\
Warmup & 10\% of steps \\
Max.\ prompt / completion & 1024 / 1024 tokens \\
Loss & forward KL, first 3 tokens masked \\
Student storage / compute & 32 / 16 bits (bfloat16) \\
Teacher storage / forward & 32 / 16 bits (bfloat16) \\
EMA rate $1-\tau$ & 0.01 \\
Grad.\ clip / weight decay & 1.0 / 0 \\
\ns{} steps $K$ / coefficients $(\kappa_1,\kappa_2,\kappa_3)$ & 5 / $(3.4445,-4.7750,2.0315)$ \\
Update scale $c_{A,B}$ & $0.2\sqrt{\max(A,B)}$ \\
\our{}: $\alpha$ / power iterations $K_{\mathrm{PI}}$ & 0.1 / 8 (warm-started) \\
AdamW (other params.) & $(0.9,0.999)$, $10^{-8}$ \\
Evaluation & greedy, every 50 steps \\
\bottomrule
\end{tabular}
\caption{\base{} training settings for all optimizers.}
\label{tab:hyperparameters}
\end{table}

Table~\ref{tab:hyperparameters} lists the \base{} settings. Following \citet{shenfeld2026sdft} and \citet{hubotter2026sdpo}, we ran every configuration with 3 random seeds. A \base{} run takes 1.1--1.8 hours on 2 H100 GPUs, dominated by generation. For \sdpo{} (Table~\ref{tab:sdpo_hyperparameters}), the model runs without thinking and every setting is identical to the experiments of \citet{hubotter2026sdpo} without feedback on Olmo-3-7B-Instruct: model, data splits, rollout and validation settings, loss, teacher EMA rate, schedule, weight decay and gradient clipping. Only the optimizer of the hidden matrices and its learning rate change. As for \base{}, Muon, MuonClip and \our{} are trained with learning rates $5{\times}10^{-5}$ and $3{\times}10^{-5}$, and we report the better run; the AdamW rows are taken from \citet{hubotter2026sdpo}. Accuracy is the best avg@16 reached within 1 and 5 hours of wall-clock training, excluding initialization and validation, as in \citet{hubotter2026sdpo}.

\begin{table}[h]
\centering
\small
\setlength{\tabcolsep}{4pt}
\begin{tabular}{@{}ll@{}}
\toprule
\textbf{Setting} & \textbf{Value} \\
\midrule
Model & Olmo-3-7B-Instruct \\
Data & SciKnowEval (2 subsets) \\
Max.\ prompt / response & 2048 / 8192 \\
Questions / mini-batch & 32 / 32 \\
Rollouts / temperature & 8 / 1.0 \\
Validation / temperature / top-$p$ & avg@16 / 0.6 / 0.95 \\
Loss / top-$k$ & Jensen--Shannon / 100 \\
EMA rate $1-\tau$ & 0.05 \\
Importance-sampling clip & 2 \\
Schedule / warmup & constant / 10 steps \\
Weight decay / grad.\ clip & 0.01 / 1.0 \\
\midrule
Student storage / compute & 32 / 16 bits (bfloat16) \\
Teacher storage / forward & 32 / 16 bits (bfloat16) \\
LR, AdamW & $1{\times}10^{-5}$ \\
LR, Muon / MuonClip / ours & $\{3,5\}{\times}10^{-5}$ \\
Muon, MuonClip & $\mu{=}0$ \\
\our{}: $\alpha$ / $K_{\mathrm{PI}}$ & 0.1 / 8 \\
\bottomrule
\end{tabular}
\caption{\sdpo{} training settings. All entries above the line are those of \citet{hubotter2026sdpo}.}
\label{tab:sdpo_hyperparameters}
\end{table}

\subsection{Memory Projection}
\label{app:projection}

Figure~\ref{fig:projection} projects the persistent model state of self-distillation for a model with $N{=}10^{13}$ parameters. Per parameter, all optimizers store the student, its gradient and the EMA-teacher in 32 bits (12 bytes); every state uses the same precision, so no optimizer is favored. AdamW adds 2 32-bit moments (8 bytes). Muon and MuonClip, in their standard form with $\mu{=}0.95$, add a 32-bit momentum (4 bytes) on hidden matrices and AdamW states on the remaining parameters; the per-head statistics of MuonClip are negligible. \our{} adds no state on hidden matrices and AdamW states on the rest. We take hidden matrices as 99\% of the parameters, since embeddings and output head take about 0.2\% of the parameters in a 1T model~\citep{kimi2025k2}. This gives 20, 16.04, 16.04 and 12.08 bytes per parameter, i.e.\ 200, 160, 160 and 121\,TB. Activations and generation buffers are excluded. Our 7B measurements on 2 GPUs, all with a 32-bit teacher, are taken between steps, when gradients are freed; without the 4-byte gradient, the same accounting gives savings of 27\% against Muon and 43\% against AdamW, close to the measured 27\% and 42\%. \our{}, momentum-free Muon and momentum-free MuonClip need 37.7\,GB per GPU. A Muon run that allocates the 32-bit momentum of 6.5B hidden parameters needs 51.6\,GB, close to $37.7+13.1=50.8$\,GB. Storing the teacher in 32 instead of 16 bits adds 7.6\,GB per GPU: Muon with $\mu{=}0.95$ and a 16-bit teacher needs 43.9\,GB, and $43.9+7.6=51.5$\,GB is close to the measurement. AdamW with a 16-bit teacher needs 56.9\,GB, so with a 32-bit teacher it needs $56.9+7.6=64.5$\,GB, the value in Figures~\ref{fig:teaser} and~\ref{fig:dynamics}(d).

\subsection{Learning-Rate Sensitivity and Training Diagnostics}
\label{app:lr}
\label{app:diagnostics}

\begin{figure}[t]
  \centering
  \includegraphics[width=\linewidth]{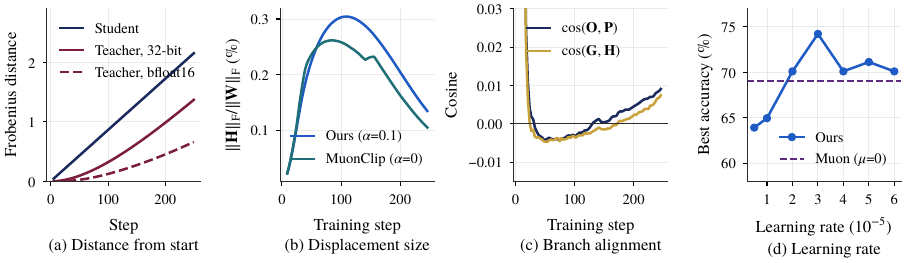}
  \caption{(a) Distance from the initial weights for the student and for teachers in 2 precisions. (b) Size of the teacher--student displacement. (c) Alignment of the 2 branches. (d) Best tool-use accuracy versus learning rate.}
  \label{fig:appendix}
\end{figure}

\noindent\textbf{Learning rate.} Figure~\ref{fig:appendix}(d) shows that \our{} exceeds the best momentum-free Muon run (69.1\%) for every learning rate from $2\times10^{-5}$ to $6\times10^{-5}$ (70.1--74.2\%) and peaks at $3\times10^{-5}$. Only at $10^{-5}$ and below do 2 epochs leave too few effective steps.

\noindent\textbf{Teacher branch.} We log, for every hidden matrix and step, $\|\mathbf{H}^{(s)}\|_{\mathrm F}/\|\mathbf{W}_S^{(s)}\|_{\mathrm F}$, $\sigma_{\max}(\mathbf{H}^{(s)})/\|\mathbf{H}^{(s)}\|_{\mathrm F}$, $\cos(\mathbf{G}^{(s)},\mathbf{H}^{(s)})$ and $\cos(\mathbf{O}^{(s)},\mathbf{P}^{(s)})$, averaged over matrices. The displacement grows during warmup, peaks at 0.2--0.3\% of the weights and shrinks as the learning rate decays (Figure~\ref{fig:appendix}(b)). Its largest singular value carries about 6\% of its Frobenius norm on all runs, so $\mathbf{H}^{(s)}$ is spread over many directions. After the first 20 steps, the 15-step averages of both cosines stay below 0.015 in magnitude on all runs (Figure~\ref{fig:appendix}(c)): the teacher branch is nearly orthogonal to the current gradient and adds directions it does not cover.

\noindent\textbf{Experiments on a Qwen2.5-7B-Instruct weight matrix.} Figure~\ref{fig:analysis}(b, c), Figure~\ref{fig:appendix}(a) and the constants below use the key projection matrix of layer 10 of Qwen2.5-7B-Instruct ($512\times3584$). The student takes momentum-free Muon steps $\mathbf{W}\gets\mathbf{W}-\eta c_{A,B}\mathrm{NS}_K(\mathbf{G}^{(s)})$ with $\eta{=}5{\times}10^{-5}$ and $\mathbf{G}^{(s)}=\mathbf{d}+\nu\,\mathbf{E}^{(s)}/\|\mathbf{E}^{(s)}\|_{\mathrm F}$, where $\mathbf{d}$ is a fixed random direction with $\|\mathbf{d}\|_{\mathrm F}{=}1$, $\mathbf{E}^{(s)}$ is fresh Gaussian noise and $\nu$ is the noise level; the teacher follows Eq.~\eqref{eq:ema_update} with $\tau{=}0.99$. For Figure~\ref{fig:analysis}(b), we run 150 steps for each $\nu$ and average, over the last 20 steps, the cosine of $\mathbf{O}^{(s)}$ and of $\mathbf{P}^{(s)}$ with $\mathrm{NS}_K(\mathbf{d})$. For Figure~\ref{fig:analysis}(c), we take $\mathbf{H}^{(s)}$ after 150 steps with $\nu{=}2$ and a gradient of low stable rank, $\mathbf{G}=\mathbf{Y}_1\mathbf{Y}_2^{\top}$ with $\mathbf{Y}_1\in\mathbb{R}^{A\times m}$, $\mathbf{Y}_2\in\mathbb{R}^{B\times m}$ Gaussian, $m\in\{4,16,64\}$, plus 1\% Gaussian noise. We compare $\|\mathbf{U}-\mathbf{O}\|_{\mathrm F}/\|\mathbf{O}\|_{\mathrm F}$ for Eq.~\eqref{eq:post_mix} and for the pre-\ns{} mix $\mathbf{U}=\mathrm{NS}_K\big((1-\alpha)\mathbf{G}/\|\mathbf{G}\|_{\mathrm F}+\alpha\mathbf{H}^{(s)}/\|\mathbf{H}^{(s)}\|_{\mathrm F}\big)$. For Figure~\ref{fig:appendix}(a), the teacher is either kept in 32 bits or rounded to bfloat16 after every EMA step.

\noindent\textbf{Constants of Theorem~\ref{thm:reuz_convergence}.} With the same matrix and gradients, the student follows \our{} ($\alpha{=}0.1$), so the teacher records \our{}'s own trajectory. Over steps 51--150 we average $a_X=\langle\mathbf{d},\mathbf{X}\rangle_{\mathrm F}$ and $b_X=\|\mathbf{X}\|_{\mathrm F}^2$ for $\mathbf{X}\in\{\mathbf{O}^{(s)},\mathbf{P}^{(s)},\mathbf{U}^{(s)}\}$, where $\mathbf{d}$ is the mean gradient. For noise levels $\nu\in\{0.5,1,2,4,8,16\}$, $a_P/a_O$ is 1.10, 1.37, 1.97, 3.20, 5.03 and 6.62, so $a_R$ exceeds $a_B{=}a_O$ by 1\%--56\%, while $b_P<b_O$ and $b_R<b_O$ (e.g.\ 366 against 447 at $\nu{=}16$). Hence $\gamma_R>\gamma_B$ for every $L\eta$. The cosine between $\mathbf{G}^{(s)}$ and $\mathbf{P}^{(s)}$ falls from 0.87 to 0.03 as $\nu$ grows, while $a_P/a_O$ rises, so a small single-batch cosine does not indicate a weak teacher direction.

\noindent\textbf{Teacher--student displacement without normalization.} For Figure~\ref{fig:analysis}(a), the predicted ratio $\|\mathbf{W}_T^{(s)}-\mathbf{W}_S^{(s)}\|_{\mathrm F}/\|\mathbf{W}_S^{(s)}\|_{\mathrm F}$ is $\tau \psi_s/\mathrm{rms}(\mathbf{W})$, where $\psi_s$ is $\sum_j\tau^j u_{s-1-j}$ (aligned updates) or $(\sum_j\tau^{2j}u_{s-1-j}^2)^{1/2}$ (uncorrelated updates), $u_s$ is the logged learning rate times the update root-mean-square, and $1/\mathrm{rms}(\mathbf{W})$ is averaged over the 196 hidden matrices. From step 20 on, the measured mean ratio is 1.2--1.7 times the uncorrelated value and 0.12--0.34 times the aligned value.

\noindent\textbf{Lemma 1 range.} For Figure~\ref{fig:analysis}(a), let $u_s=\eta_s\rho_s$ with the logged learning rate $\eta_s$ and update root-mean-square $\rho_s$. By Eq.~\eqref{eq:trajectory_history} and the triangle inequality, $\|\mathbf{W}_T^{(s)}-\mathbf{W}_S^{(s)}\|_{\mathrm F}$ is proportional to at most $\sum_j\tau^j u_{s-1-j}$ (aligned updates) and, for uncorrelated updates, to $(\sum_j\tau^{2j}u_{s-1-j}^2)^{1/2}$. We normalize both and the measured $\|\mathbf{H}^{(s)}\|_{\mathrm F}/\|\mathbf{W}_S^{(s)}\|_{\mathrm F}$ by their maxima.

\subsection{Attention Logits and QK-Clip}
\label{app:qk}

Growing attention logits can destabilize training with Muon, which motivated QK-Clip~\citep{kimi2025k2,wortsman2024proxies,dehghani2023vit22b}. We measured the maximum pre-softmax logit of every head (after rotary embeddings) on 24 evaluation prompts. The base Qwen2.5-7B-Instruct already reaches about 4{,}800, and 26--28 heads exceed the threshold of 100 used by \citet{kimi2025k2}; these are attention-sink heads driven by the query and key biases~\citep{xiao2024sinks}. After \base{}, the global maximum stays within $\pm1\%$ of the base value for AdamW, Muon and \our{} alike, the largest heads grow by at most 13--20\%, and the norms of the query and key weights change by less than 0.05\%. Clipping at 100 would therefore rescale about 27 heads that are large by design. Our MuonClip baseline instead uses per-head thresholds of $1.1\times$ the largest logit observed in the first 10 steps. Because logits do not grow, QK-Clip is not needed in this setting. With \our{}, it is also harmful (Table~\ref{tab:ablation}(d)). In the tool-use run with clipping, 8 clips occur, each on 1 head, with factors down to 0.56; right after the first clip (step 20, factor 0.90), the mean $\|\mathbf{H}^{(s)}\|_{\mathrm F}/\|\mathbf{W}_S^{(s)}\|_{\mathrm F}$ over all hidden matrices rises 12\% above the run without clipping, after tracking it within 2\% before. To isolate the effect, we take the query matrix of layer 10 of Qwen2.5-7B-Instruct, scale the 128 rows of 1 head in the student but not in the teacher, and form $\mathbf{H}^{(s)}$ and $\mathbf{P}^{(s)}$ as in Eqs.~\eqref{eq:radial_projection}--\eqref{eq:spectral_normalization}. A factor of 0.90 (0.56) creates a gap of 1.9\% (8.0\%) of the matrix norm, against 0.1--0.3\% from learning; radial removal changes it by under 1.5\%, 97\% (99\%) of $\|\mathbf{H}^{(s)}\|_{\mathrm F}^2$ lies in the clipped head, and $-\mathbf{P}^{(s)}$ restricted to that head has cosine $-1.00$ with its weights. The update therefore shrinks the clipped head further at every step. The offset decays by $\tau$ per step, so a 1.9\% gap falls below 0.2\% only after about 225 steps. Scaling the same query rows of the teacher by the same factor at every clip keeps $\mathbf{W}_T^{(s)}-\mathbf{W}_S^{(s)}$ free of this offset; in Table~\ref{tab:ablation}(d), this variant reaches 67.3 on tool use and 70.1 on science.

\subsection{Training and evaluation.}
\label{sec:app_train_eval}

Only the optimizer changes; each host method keeps its loss, batch size and EMA settings same. For every optimizer, the student is trained in bfloat16 mixed precision (bfloat16 compute, 32-bit master weights), and the EMA-teacher is stored in 32 bits with its forward pass in bfloat16. Appendix~\ref{app:fp32} shows that this precision does not affect accuracy. We train Muon, MuonClip and \our{} ($\alpha{=}0.1$) with learning rates $5{\times}10^{-5}$ and $3{\times}10^{-5}$ and report the better run. Following \citet{shenfeld2026sdft}, we report the best checkpoint under greedy decoding and measure prior capabilities on HellaSwag, HumanEval, IFEval, MMLU, TruthfulQA, and Winogrande. For \sdpo{}, we report the best avg@16 within 1 and 5 hours of training~\citep{hubotter2026sdpo}.

\subsection{Empirical Checks of the Theory}
\label{sec:app_exp_theory}

Figure~\ref{fig:analysis} tests each theoretical statement with a matching measurement.

\noindent\textbf{Lemma~\ref{lem:ema_history}: the teacher stores applied updates.} Eq.~\eqref{eq:trajectory_history} predicts the size of $\mathbf{W}_T^{(s)}-\mathbf{W}_S^{(s)}$ from the learning rate and the size of the applied updates alone: it lies between the sum of their norms (fully aligned updates) and the root of the sum of their squares (uncorrelated updates). The measured displacement follows this range closely, with correlation 0.97 (Figure~\ref{fig:analysis}(a)). It peaks near step 107, after the learning rate peaks, exactly as an EMA of applied updates should. Without normalization, the measured size lies inside this range at every step after step 20, at 1.2--1.7 times the uncorrelated bound, so the applied updates share a persistent component (Appendix~\ref{app:diagnostics}).

\noindent\textbf{Lemma~\ref{lem:ema_variance}: the teacher averages out noise.} On a Qwen2.5-7B-Instruct weight matrix trained with noisy gradients of known mean, the current orthogonalized gradient loses the mean direction as gradient noise grows (cosine 0.87 to 0.11), while the teacher branch keeps it (0.99 to 0.75; Figure~\ref{fig:analysis}(b)).

\noindent\textbf{Proposition~\ref{prop:polar_alignment}: the gain is temporal.} During training, the 2 branches are nearly orthogonal, with $|\cos(\mathbf{O}^{(s)},\mathbf{P}^{(s)})|<0.015$ after the first steps (Appendix~\ref{app:diagnostics}). The teacher branch thus adds directions that the current gradient does not contain, as the proposition requires.

\noindent\textbf{Post-\ns{} mixing bounds the teacher.} Mixed before \ns{}, even $\alpha{=}0.02$ changes the update by more than its own norm, because \ns{} lifts the teacher's weak directions to unit singular value (Figure~\ref{fig:analysis}(c)). Mixed after \ns{}, as in \our{}, the change grows linearly with $\alpha$, as Eq.~\eqref{eq:post_mix} guarantees.

\noindent\textbf{Theorem~\ref{thm:reuz_convergence} and Corollary~\ref{cor:faster_than_baseline}: faster convergence.} Under the PL condition, a smaller gradient norm means a smaller optimality gap. At the same learning rate, \our{} has the lowest gradient norm from step 40 onward (Figure~\ref{fig:analysis}(d)) and the lowest self-distillation KL for most of training (Figure~\ref{fig:dynamics}(c)), and it reaches its best accuracy earlier (Figure~\ref{fig:dynamics}(a)). These observations are consistent with the corollary. We also check its condition directly by estimating the constants of A4 on the Qwen2.5-7B-Instruct weight matrix with gradients of known mean (Appendix~\ref{app:diagnostics}): at every noise level, the teacher branch is better aligned with the mean gradient than the orthogonalized gradient (up to 6.6 times) and has a smaller second moment, so $a_R>a_B$ and $b_R<b_B$, which gives $\gamma_R>\gamma_B$ for any step size. The single-batch cosine between gradient and teacher direction stays small (0.03 at the highest noise), so the near-zero cosines measured in training do not contradict this condition.

\subsection{Ablations and Analysis}
\label{sec:app_exp_ablation}

We ablate each component of \our{} with \base{} on both tasks (Table~\ref{tab:ablation}): the optimizer state it replaces, the teacher weight $\alpha$, the construction of the teacher branch, and its combination with logit clipping. All ablations follow the setting of Table~\ref{tab:continual_overall}, with $\alpha{=}0.1$ unless varied, and every block includes the full method as reference.

\noindent\textbf{Which history helps.} Gradient history and teacher history have opposite effects (Table~\ref{tab:ablation}(a)). Standard momentum reduces accuracy by 0.7--6.2 points for both Muon and MuonClip on both tasks, whereas the teacher branch raises the accuracy of momentum-free Muon by 5.1 and 2.8 points. Orthogonalization normalizes each update, but with momentum consecutive updates share a direction for about $1/(1-\mu){=}20$ steps and add up instead of partly cancelling, so the weights move further at the same learning rate. In addition, self-distillation targets move with the teacher at every step, so an average of past gradients points partly toward targets that no longer apply. The teacher branch, by contrast, records the updates that were actually applied (Lemma~\ref{lem:ema_history}), and the bounded mix adds this history without letting it override the current gradient.

\noindent\textbf{Teacher weight.} $\alpha{=}0.1$ is the best value on both tasks (Table~\ref{tab:ablation}(b)). Both smaller and larger values reduce accuracy, confirming that the teacher should guide, not replace, the gradient direction (Eq.~\eqref{eq:post_mix}). The same weight is optimal for both tasks and is used without modification for \sdpo{}, so $\alpha$ requires no per-task tuning.

\noindent\textbf{Teacher branch.} Every modification to the teacher branch reduces accuracy (Table~\ref{tab:ablation}(c)), by 1.7--2.6 points for mixing before \ns{} up to 5.0 points for a cosine gate. The gradient and displacement are nearly orthogonal ($|\cos|\approx0.004$), causing the cosine gate to effectively switch the teacher off. On the tool-use task, even the weaker variants outperforms the momentum-free Muon (70.5--72.5 Vs. 69.1), indicating that the teacher's history is beneficial in every form; however, across both tasks, the bounded post-\ns{} formulation extracts the most from it.

\noindent\textbf{QK-Clip.} Logit clipping helps Muon (69.1 to 73.2 on tool use) but hurts \our{} (Table~\ref{tab:ablation}(d); 71.1 to 66.0 at the same learning rate). QK-Clip rescales a head's query weights in the student only, so each clip introduces an offset into $\mathbf{W}_T^{(s)}-\mathbf{W}_S^{(s)}$ that is not learning motion. On a real query matrix, a single clip by a factor 0.9 creates a gap 6--19 times larger than the learning displacement, this gap is retained by radial removal and accounts for 97\% of $\mathbf{H}^{(s)}$ (Appendix~\ref{app:qk}). As a result, the teacher branch continues shrinking the clipped head at full weight $\alpha$ for over 200 steps; in training, the first clip raises the mean displacement over all matrices by 12\% within a single step. Applying the same clip to the teacher removes this offset and recovers 1.3 points on tool use (67.3), but both tasks remain below \our{} without clipping (67.3 and 70.1): each clip also perturbs the teacher's attention patterns and therefore the self-distillation target. In any case, clipping is unnecessary here, as logits stay within 1\% of their initial maximum for every optimizer.